%% file: main.tex
\pdfoutput=1
\documentclass[10pt,twocolumn,letterpaper]{article}

\usepackage[pagenumbers]{cvpr} %
\usepackage[accsupp]{axessibility}

\input{shortcut}

\def\paperID{1800} %
\def\confName{CVPR}
\def\confYear{2025}

\begin{document}
\title{Sonata: Self-Supervised Learning of Reliable Point Representations}

\author{Xiaoyang Wu\textsuperscript{\mdseries1,2} \authorskip 
Daniel DeTone\textsuperscript{\mdseries2*} \authorskip
Duncan Frost\textsuperscript{\mdseries2*} \authorskip
Tianwei Shen\textsuperscript{\mdseries2*} \authorskip
Chris Xie\textsuperscript{\mdseries2*} \authorskip
Nan Yang\textsuperscript{\mdseries2*} \\
Jakob Engel\textsuperscript{\mdseries2} \authorskip
Richard Newcombe\textsuperscript{\mdseries2} \authorskip
Hengshuang Zhao\textsuperscript{\mdseries1} \authorskip
Julian Straub\textsuperscript{\mdseries2}
 \\ \\
\textsuperscript{1}The University of Hong Kong \institutionskip
\textsuperscript{2}Meta Reality Labs Research \\ 
{\tt\small \url{https://github.com/facebookresearch/sonata}}
}

\twocolumn[{%
\renewcommand\twocolumn[1][]{#1}%
\vspace{-17mm}
\maketitle
\vspace{-11mm}
\begin{center}
    \captionsetup{type=figure}
    \includegraphics[width=\linewidth]{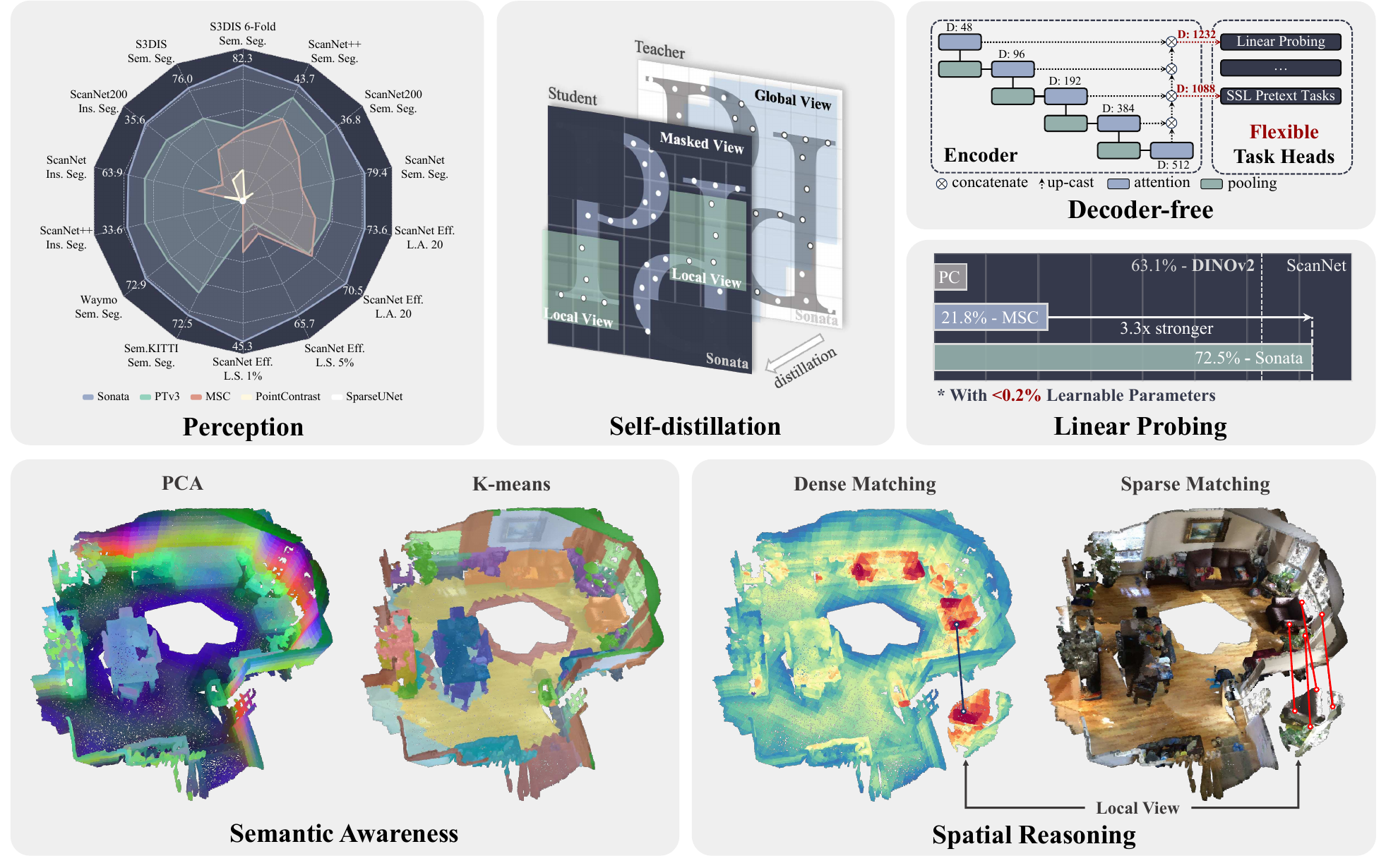}
    \vspace{-7mm}
    \captionof{figure}{
    \textbf{Main properties.}
    Sonata leads to reliable 3D self-supervised pretraining with the following superior and emerging properties: 1.~\textbf{\textit{Perception}}. Sonata advances state-of-the-art results across 3D indoor and outdoor perception tasks; 2.~\textbf{\textit{Linear probing}}. With less than 0.2\% learnable parameters, Sonata achieves strong and usable linear probing performance which is 3.3$\times$ better than previous SOTA; 
    3.~\textbf{\textit{Decoder-free}}. Sonata moves beyond the inflexible U-Net structure, offering multi-scale representations that unchain future 3D research from previous architectural constraints. 
    4.~\textbf{\textit{Semantic awareness}}.  Sonata reveals semantic structure in PCA and K-means visualizations. 5.~\textbf{\textit{Spatial reasoning}}. Sonata allows spatial correspondence even under strong augmentations as visualized via feature similarity. 
    }\label{fig:teaser}
    \vspace{-2mm}
\end{center}%
}]

\begin{abstract}
    \input{./section/0_abstract}
\end{abstract}

\input{./section/1_introduction}

\input{./section/2_related}

\input{./section/3_method}

\input{./section/4_experiments}

\input{./section/5_conclusion}

\section*{Acknowledgements}
We extend our gratitude to Maxime Oquab and Piotr Bojanowski for their guidance on the DINOv2 training details, to Saining Xie for insightful discussions on the vision of 3D representation learning, and to Paul Mcvay for his thoughts on the JPEA framework with sparse point cloud data.

\appendix
\section*{Appendix}
\input{./section/6_appendix}

{
\small
\bibliographystyle{ieeenat_fullname}
\bibliography{main}
}

\end{document}

%% file: shortcut.tex
\usepackage{lipsum}
\usepackage{array}
\usepackage{times}
\usepackage{epsfig}
\usepackage{graphicx}
\usepackage{float}
\usepackage{wrapfig}
\usepackage{amsmath,amssymb,amsthm}
\usepackage{algorithm,algorithmicx,algpseudocode}
\usepackage{bm,xspace}
\usepackage{comment}
\usepackage{multirow}
\usepackage{balance}
\usepackage{url}
\usepackage{booktabs}
\usepackage{etoolbox,siunitx}
\usepackage{calc}
\usepackage{pifont,hologo}
\usepackage{color}
\usepackage{adjustbox}
\usepackage{amsmath}
\usepackage{enumitem}
\usepackage{bbding}  %
\usepackage[normalem]{ulem}  %
\usepackage{contour}
\usepackage[table]{xcolor}
\usepackage{soul}%
\usepackage{tocloft} %
\usepackage{etoc} %
\PassOptionsToPackage{dvipsnames}{xcolor}
\usepackage{algorithm}
\usepackage{algpseudocode}
\usepackage{listings}
\usepackage[pagebackref,breaklinks,colorlinks,linkcolor=link,citecolor=cite,breaklinks=true]{hyperref}
\usepackage{natbib}
\PassOptionsToPackage{square,numbers,comma,sort}{natbib}
\usepackage[switch]{lineno}

\newcommand{\authorskip}{\hspace{2.5mm}}
\newcommand{\institutionskip}{\hspace{5.0mm}}

\definecolor{darkblue}{HTML}{35394B}
\definecolor{blue}{HTML}{004bb3}
\definecolor{red}{HTML}{cc1100}
\definecolor{orange}{HTML}{cc7700}
\definecolor{gray}{HTML}{efefef}
\definecolor{darkgreen}{HTML}{228B22}
\definecolor{darkgray}{HTML}{808080}
\definecolor{lightpurple}{HTML}{a56dba}

\definecolor{cite}{HTML}{3270b5}
\definecolor{link}{HTML}{b53532}
\definecolor{link}{HTML}{cc1100}
\definecolor{scratch}{HTML}{001219}
\definecolor{pretrain}{HTML}{0a9396}

\newcommand{\scratch}{\textcolor{scratch}{$\mathbf{\circ}$\,}}
\newcommand{\pretrain}{\textcolor{pretrain}{$\bullet$\,}}
\newcommand{\plus}{\textcolor{darkblue}{\raisebox{0.3ex}{\tiny +\,}}}

\contourlength{0.8pt}

\newcommand{\figref}[1]{Fig.~\ref{#1}}
\newcommand{\tabref}[1]{Tab.~\ref{#1}}
\newcommand{\secref}[1]{Sec.~\ref{#1}}
\renewcommand{\eqref}[1]{Eq.~\ref{#1}}

\newcolumntype{x}[1]{>{\centering\arraybackslash}p{#1}}
\newcolumntype{y}[1]{>{\raggedright\arraybackslash}p{#1}}
\newcolumntype{z}[1]{>{\raggedleft\arraybackslash}p{#1}}
\newcommand{\tablestyle}[2]{\setlength{\tabcolsep}{#1}\renewcommand{\arraystretch}{#2}\centering\footnotesize}

\DeclareMathSymbol{@}{\mathord}{letters}{"3B}

\newcommand\mypara[1]{\vspace{0mm}\noindent\textbf{#1}}

\newcommand\tinyless{\raisebox{0.3ex}{\tiny $<$}}

\makeatletter
\DeclareRobustCommand\onedot{\futurelet\@let@token\@onedot}
\def\@onedot{\ifx\@let@token.\else.\null\fi\xspace}
\def\eg{\emph{e.g}\onedot} 
\def\ie{\emph{i.e}\onedot}

\newcommand*{\Rom}[1]{\expandafter\@slowromancap\romannumeral #1@}
\newcommand*{\rom}[1]{\expandafter\romannumeral #1}

\def\1{\bm{1}}

\DeclareMathAlphabet{\mathsfit}{\encodingdefault}{\sfdefault}{m}{sl}
\SetMathAlphabet{\mathsfit}{bold}{\encodingdefault}{\sfdefault}{bx}{n}

\let\originalleft\left
\let\originalright\right
\renewcommand{\left}{\mathopen{}\mathclose\bgroup\originalleft}
\renewcommand{\right}{\aftergroup\egroup\originalright}

%% file: section/0_abstract.tex
\begin{nolinenumbers}
\\
\vspace{-12mm}
\end{nolinenumbers}
\renewcommand{\thefootnote}{\fnsymbol{footnote}}
\footnotetext[1]{Equal contribution in alphabetic order.}

In this paper, we question whether we have a reliable self-supervised point cloud model that can be used for diverse 3D tasks via simple linear probing, even with limited data and minimal computation. We find that existing 3D self-supervised learning approaches fall short when evaluated on representation quality through linear probing. We hypothesize that this is due to what we term the ``geometric shortcut'', which causes representations to collapse to low-level spatial features. This challenge is unique to 3D and arises from the sparse nature of point cloud data. We address it through two key strategies: obscuring spatial information and enhancing the reliance on input features, ultimately composing a \textbf{Sonata} of 140k point clouds through self-distillation. Sonata is simple and intuitive, yet its learned representations are strong and reliable: zero-shot visualizations demonstrate semantic grouping, alongside strong spatial reasoning through nearest-neighbor relationships. Sonata demonstrates exceptional parameter and data efficiency, tripling linear probing accuracy (from 21.8\% to 72.5\%) on ScanNet and nearly doubling performance with only 1\% of the data compared to previous approaches. Full fine-tuning further advances SOTA across both 3D indoor and outdoor perception tasks.\looseness=-1
\clearpage

%% file: section/1_introduction.tex
\vspace{-13.5mm}
\section{Introduction}
\vspace{-1mm}
Self-supervised learning (SSL) with images~\cite{chen2020simclr,he2020moco,zhang2022dino,assran2023jpea, xie2022simmim} has seen a continuous increase in model simplicity, capacity, and capability over the past decade~\cite{doersch2015unsupervised}. Tuning a single linear layer can achieve performance close to full fine-tuning~\cite{zhang2022dino,grill2020byol,he2022mae}, fostering growing trust in its reliability. This trust has been further strengthened by witnessing the semantic meaning of learned image representations through direct visualization~\cite{zhou2021ibot,oquab2023dinov2}. 
Consequently, these reliable self-supervised models have become the foundation for emerging approaches~\cite{yang2024depthanything,yang2024depthanythingv2,wang2024dust3r,chen2024anydoor} across various fields involving images. \looseness=-1

In contrast to the image domain, self-supervised learning with point clouds~\cite{choy2019fully,xie2020pointcontrast,Pang2022pointmae,yu2021pointbert,zhang2022pointm2ae} is still in its early stages. Despite the broad reliance on 3D applications in autonomous driving~\cite{sun2020waymo,caesar2020nuscenes}, robotic learning~\cite{gu2023maniskill2}, mixed reality~\cite{dehghan2021arkitscenes,james2019rlbench} and egocentric perception~\cite{engel2023aria, straub2024efm3d}, the latest self-supervised point cloud models are seldom included in their pipelines. This gap prompts us to consider a simple yet critical question: \textit{do we have a reliable point self-supervised learning approach} that provides strong representations, usable with simple linear probing across these applications? Not yet. Previous SOTAs~\cite{xie2020pointcontrast,hou2021csc,wu2023msc} fall short on this higher-level criterion, achieving only a maximum of 21.8\% mIoU on ScanNet~\cite{dai2017scannet} with linear probing, especially given that the performance from scratch is 77.6\%.

\begin{figure}[t]
    \vspace{-0mm}
    \centering
    \includegraphics[width=\linewidth]{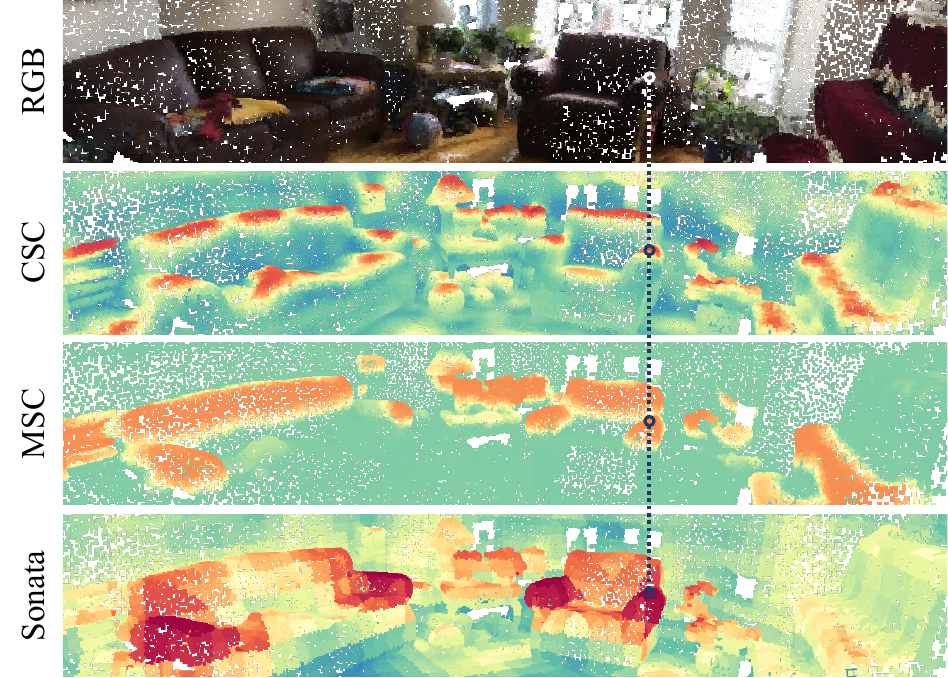}
    \vspace{-7.5mm}
    \caption{\textbf{Geometric shortcut.} We select a point on the sofa arm and compute pairwise similarity with other points. The similarity heatmap reveals that CSC~\cite{hou2021csc} collapses to surface normals, and MSC~\cite{wu2023msc} overfits to point height. In contrast, our Sonata can extract higher-level concepts, as can be seen by the high similarity between all sofa arms highlighted in \textcolor{red}{red}.}
    \label{fig:geometric_shortcut}
    \vspace{-6mm}
\end{figure}

We identify the \textit{geometric shortcut} as the primary issue hindering these prior point cloud SSL approaches from learning reliable representations. This shortcut refers to the tendency of the model to collapse to easily accessible, low-level geometric cues, such as normal direction or point height, as demonstrated in prior works and visualized in~\figref{fig:geometric_shortcut}. This spatial information is inevitably introduced into point cloud operators along with point coordinates rather than through input features, making it difficult to obscure and nearly impossible to mask effectively.

However, the model collapse caused by geometric shortcuts can be mitigated through two key strategies: \textit{obscuring spatial information} and \textit{emphasizing input features}. Specifically, we address this issue by applying SSL losses at coarser spatial scales, disturbing the spatial information of masked points without features, and progressively increasing task difficulty to reduce reliance on accessible geometric cues. Coupled with a point self-distillation framework and scaling techniques inspired by recent advances in image SSL~\cite{he2020moco,caron2020swav,zhou2021ibot,caron2021emerging,oquab2023dinov2}, we ultimately compose a \textbf{Sonata} using 140k point cloud scenes~\cite{dai2017scannet,yeshwanth2023scannet++,armeni2016s3dis,dehghan2021arkitscenes,ramakrishnan2021hm3d,zheng2020structured3d,avetisyan2024scenescript}.

Sonata demonstrates strong zero-shot capabilities, with PCA-colored visualizations of point clouds, k-means clustering of features, and nearest-neighbor matching between point clouds (see~\figref{fig:teaser}).  Sonata also proves highly data-efficient, raising semantic segmentation performance from 25.8\% to 45.3\% under extremely limited data conditions (1\% of ScanNet). Additionally, Sonata significantly boosts linear probing accuracy on ScanNet semantic segmentation, increasing it by over 3.3$\times$ from 21.8\% to 72.5\% and surpassing the accuracy of DINOv2 features aggregated onto the point cloud (63.1\%). Moreover, combining Sonata features with DINOv2 features further enhances accuracy (76.4\%), underscoring that Sonata captures unique 3D information beyond what is visible in images alone. Finally, Sonata achieves state-of-the-art results across various indoor and outdoor perception tasks with full fine-tuning.

%% file: section/2_related.tex
\vspace{-1mm}
\section{Related Work}
\vspace{-1mm}
\mypara{Image self-supervised learning.} Over the past decade, remarkable advancements~\cite{doersch2015unsupervised,noroozi2016unsupervised,wu2018unsupervised,dosovitskiy2015discriminative,hjelm2018learning} have been made in image self-supervised learning, and our research is largely inspired by two pivotal moments in this field. First, linear probing, a method that assesses representation quality by optimizing a minimal linear transformation, has become a standard in 2D image SSL~\cite{caron2018deep,wu2018unsupervised,he2020moco,chen2020simclr}. In some cases, such as when the distribution shift is large, linear probing surpasses full fine-tuning~\cite{zhang2022fine}. Second, the ability to directly perceive the semantic meaning of learned representations through zero-shot visualization like PCA or attention~\cite{caron2021emerging,oquab2023dinov2} has further strengthened trust in reliability.

\mypara{Point self-supervised learning.} Sonata follows the research path initiated by PointContrast~\cite{xie2020pointcontrast,hou2021csc}, emphasizing self-supervised learning with scene-level data~\cite{dai2017scannet}. While previous efforts do implement strategies to prevent collapse from geometric shortcuts, they remain limited. For example, Masked Scene Contrast (MSC)~\cite{wu2023msc} encourages learning beyond naive geometric cues by predicting color or normal vectors but still partially anchors representations to predefined tasks. GroupContrast (GC)~\cite{wang2024gc} employs graph-based segment guidance~\cite{felzenszwalb2004efficient}, though it is constrained by its reliance on human-designed algorithms. Building on MSC, Sonata directly addresses the geometric shortcut and scales up training~\cite{wu2024ppt} to establish a more reliable approach to point cloud self-supervised learning.

\mypara{Point cloud backbones.} This area~\cite{thomas2019kpconv,choy20194d,liang2024pointmamba,peng2024oacnns,ma2022rethinking,qian2022pointnext} has significantly benefited from the U-Net structure~\cite{ronneberger2015unet}, particularly with its hierarchical decoding and skip connections, as first introduced by PointNet++~\cite{qi2017pointnet++}. However, the tight coupling between the encoder and decoder restricts flexibility and generalization capacity~\cite{xie2021segformer}. Sonata addresses this limitation by focusing self-supervised learning exclusively on the encoder, thus removing the hierarchical decoder. Moreover, unlike previous SOTAs~\cite{xie2020pointcontrast,hou2021csc,wu2023msc} in point SSL, which primarily use SparseUNet~\cite{SubmanifoldSparseConvNet,choy20194d,spconv2022} to balance efficiency and accuracy, we leverage Point Transformer V3 (PTv3)~\cite{wu2024ptv3}, an efficient, accurate, and scalable transformer backbone. This shift alone yields a 7.7\% improvement in linear probing performance over MSC~\cite{wu2023msc}.

%% file: section/3_method.tex
\vspace{-1mm}
\section{Pilot Study and Design Principle}
\label{sec:pilot&principles}
\vspace{-1mm}
In this section, we qualitatively study the problems of current point self-supervised approaches and point cloud backbones to inform the Sonata approach.

\begin{figure}[t]
    \vspace{-2mm}
    \centering
    \includegraphics[width=\linewidth]{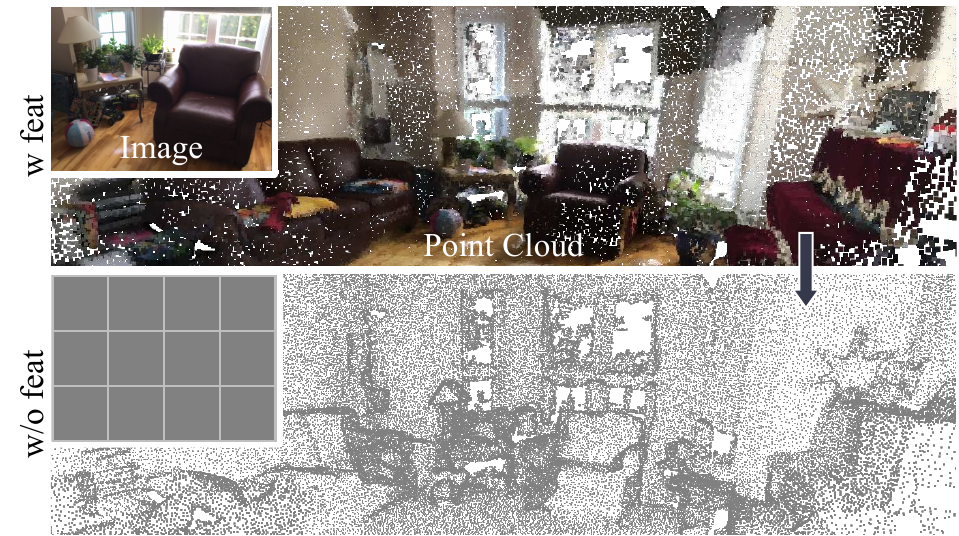}
    \vspace{-7mm}
    \caption{\textbf{The geometric shortcut is unique to 3D.} When comparing the information contained in 2D image and 3D point cloud data after removing the input feature (indicated by color), it is evident that in images all information is within the input feature. Whereas point clouds retain geometric information in point positions, which is directly utilized by operators. This characteristic leads to what we term geometric shortcuts in 3D SSL.}
    \label{fig:unique}
    \vspace{-6mm}
\end{figure}

\mypara{Uncovering the geometric shortcut in point SSL.}
The history of image self-supervised learning~\cite{zhou2022simple,he2020moco,zhang2022dino,oquab2023dinov2} can be summarized as a continuous battle against shortcuts, where models often exploit trivial solutions (mode collapse) rather than understanding deeper semantics. Each advancement has involved identifying, understanding, and overcoming these shortcuts, refining pretext tasks to push models to ``struggle'' in learning stronger representations.

However in 3D point SSL, despite various attempts~\cite{xie2020pointcontrast,hou2021csc,wu2023msc} to increase the difficulty of pretext tasks, a curious phenomenon persists---the loss consistently reduces rapidly to an ideal range in the early stages of training. We hypothesize that this lack of ``struggle'' in learning indicates that a shortcut exists that leads to a collapse of the representation to trivial solutions. Indeed as~\figref{fig:geometric_shortcut} illustrates, previous 3D SSL approaches seem to learn representations that are sensitive to local surface normals or point height. The collapse to such naive solutions even for well-designed SSL pretext tasks is what we term the \textit{geometric shortcut}.

Intuitively, representations affected by this geometric shortcut have not learned sufficient semantics. To quantify this problem, we leverage the standard criteria from 2D SSL of linear probing: we linearly probe the semantic class of each point from the learned representation.   
As illustrated in \figref{fig:teaser}, with linear probing, previous 3D SSL methods PC~\cite{xie2020pointcontrast} and MSC~\cite{wu2023msc} indeed achieve only 5.6\% mIoU and 21.8\% mIoU respectively on ScanNet semantic segmentation. When compared to the 63.1\% mIoU achieved by 3D-aggregated image representations from DINOv2~\cite{oquab2023dinov2}, it becomes obvious that the current 3D SSL methods do not learn semantic information. 

We hypothesize that the geometric shortcut stems from the sparse nature of point cloud data. Every operator, whether for point clouds or images, inherently relies on point (pixel) coordinates to define the kernel. However, unlike images with regularly spaced dense pixels, the sparsity of point cloud data 
necessitates introducing point coordinate information into point cloud operators 
rather than just through input features (see \figref{fig:unique}). 
This fact makes it difficult to obscure and nearly impossible to effectively mask out, ultimately resulting in the geometric shortcut.

These observations motivate the training methodology of Sonata aimed to prevent the dominance of naive spatial information inherent in point cloud organization.

\begin{figure}[t]
    \vspace{-2mm}
    \centering
    \includegraphics[width=\linewidth]{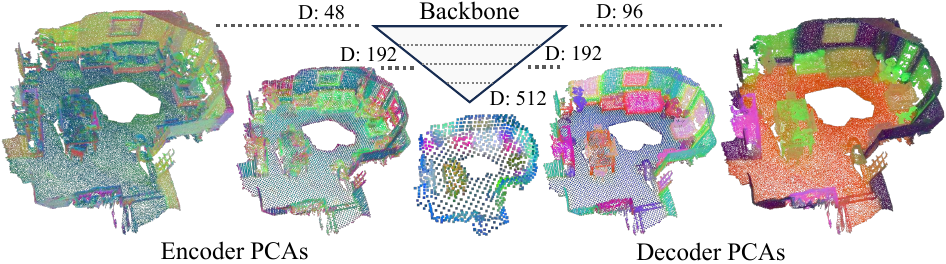}
    \vspace{-7mm}
\caption{\textbf{What is learned by the hierarchical backbone?} We visualize PCA embeddings from different stages of a hierarchical encoder and decoder, trained for \textit{semantic segmentation}. The encoder captures diverse and dispersed feature patterns, indicating a broad range of information. Notably, as the point cloud becomes coarser, accessible geometric information within point coordinates becomes increasingly global. In contrast, the decoder’s representations are more uniform and structured, suggesting a focus on refining features for task-specific outputs. \looseness=-1}
    \label{fig:hierarchical_pca}
    \vspace{-6mm}
\end{figure}

\mypara{Focusing self-supervised learning on encoder only.}
Following our hypothesis that the geometric arrangement leaks information into the learned representation via the geometric shortcut, we examine the point cloud model. The typical U-Net structure~\cite{ronneberger2015unet,qi2017pointnet++} is effective at handling large point clouds in a coarse-to-fine way, but the tight coupling of the encoder and decoder via skip connections restricts flexibility. Specifically, the decoder enforces per-point features at the original high-resolution scale, with shallow feature channels.
This constraint limits the capacity to provide richer representations, which is crucial for SSL as evidenced by large channel dimensions in state-of-the-art 2D approaches~\cite{caron2021emerging}. Most importantly, decoding point clouds at the original scale unavoidably introduces local geometric cues into operators facilitating the geometric shortcut. 

In \figref{fig:hierarchical_pca}, we visualize PCA embeddings from different stages of PTv3 encoder and decoder, trained in a supervised manner for semantic segmentation. 
We observe that the encoder learns diverse features capturing spatial features at different scales in contrast to the decoder which produces task-specific higher-level representations. 
Additionally, as the spatial resolution of the point cloud decreases via the max-pooling stages, feature representations become less local. This resolution reduction fundamentally limits reliance on fine spatial information tied to point coordinates.

These observations motivate us to remove the decoder during self-supervised learning for two main reasons. First and foremost, training directly with features at coarser point resolutions inherently restricts access to fine-grained spatial information, reducing the possibility of geometric shortcuts. Second, task-specific features can be probed or fine-tuned on top of a more expressive multi-scale point representation.

\vspace{-1mm}
\section{Point Self-distillation with Sonata}
\begin{figure}[t]
    \vspace{-2mm}
    \centering
    \includegraphics[width=\linewidth]{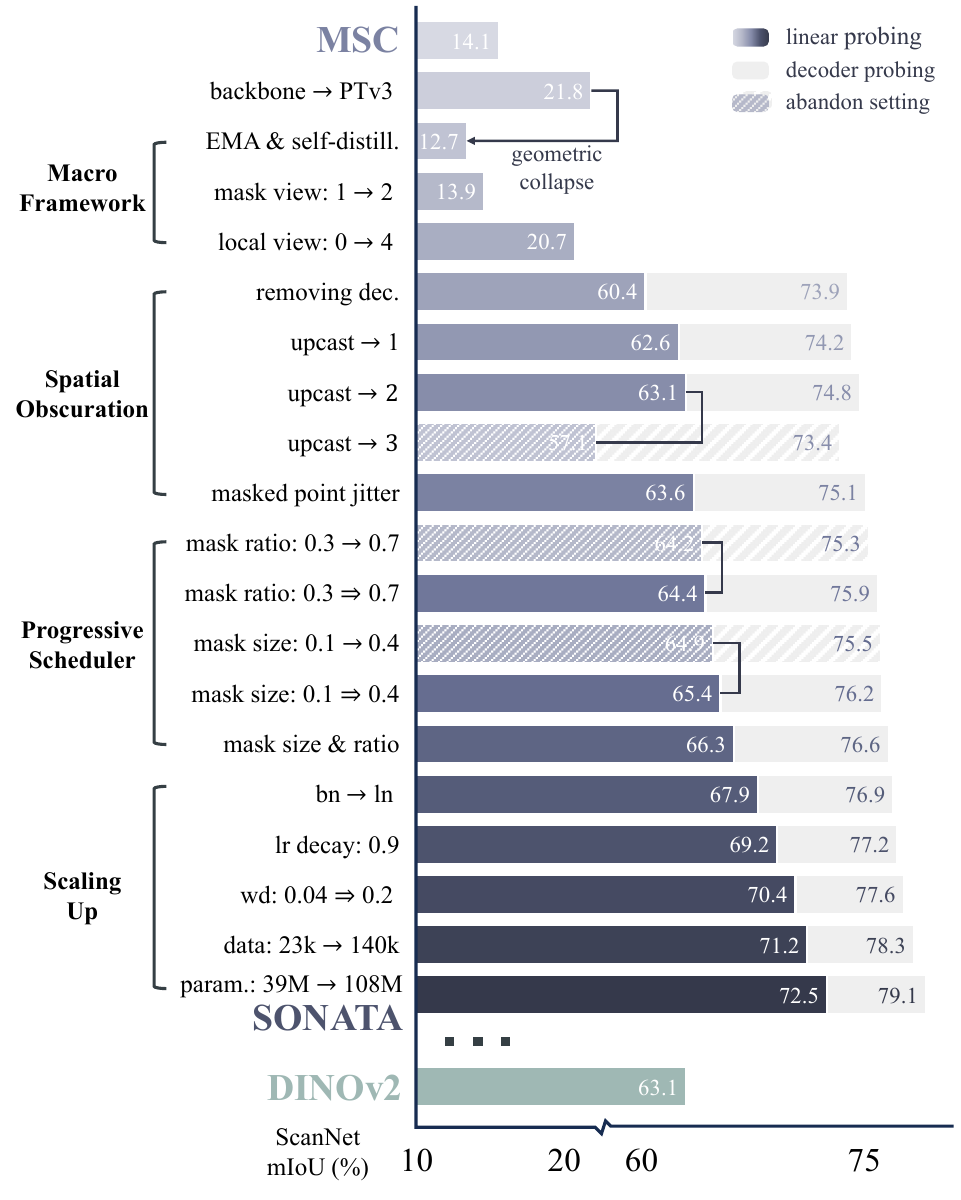}
    \vspace{-8mm}
\caption{\textbf{The roadmap.} We evolve Mask Scene Contrast~\cite{wu2023msc} into our Sonata by modernizing self-supervised learning with self-distillation, addressing the geometric shortcut, and scaling up training. Our designs are validated through progressive ablation with linear and decoder probing on ScanNet semantic segmentation~\cite{dai2017scannet}. Starting with 23k training data (a combination of ScanNet and Structured3D~\cite{zheng2020structured3d}) and a 39M PTv3 model~\cite{wu2024ptv3}, we ultimately scale up to 140k assets (\tabref{tab:data_collection}) and a 108M PTv3 model.}
    \label{fig:roadmap}
    \vspace{-4mm}
\end{figure}

\vspace{-1mm}
This section details the methodology of Sonata, the point self-distillation framework designed to address geometric shortcuts, remove structural constraints, and deliver strong linear probing results as discussed in \secref{sec:pilot&principles}. A roadmap of incremental ablation is illustrated in \figref{fig:roadmap}. 

\vspace{-2mm}
\subsection{Macro Framework}
\vspace{-2mm}
Before diving into the specific micro designs, we begin with a point self-distillation framework derived from insights gained through previous efforts in point~\cite{xie2020pointcontrast, wu2023msc} and image~\cite{he2020moco,caron2020swav,zhou2021ibot,zhang2022dino,oquab2023dinov2} self-supervised learning. This macro framework provides a solid foundation for pretext task design, allowing us to focus on addressing the geometric shortcut without additional concerns.

In essence, (point) self-supervised learning aims to \textit{make things (points) that should be the same, the same (identical in representation)}. This forms the basic recipe of point SSL: generating two views of a given point cloud with random spatial (e.g., crop, rotate, distort) and photometric (e.g., jitter) augmentations, then matching and aligning the feature embeddings of points that are close in the original space.

\begin{figure}[t]
    \vspace{-2mm}
    \centering
    \includegraphics[width=\linewidth]{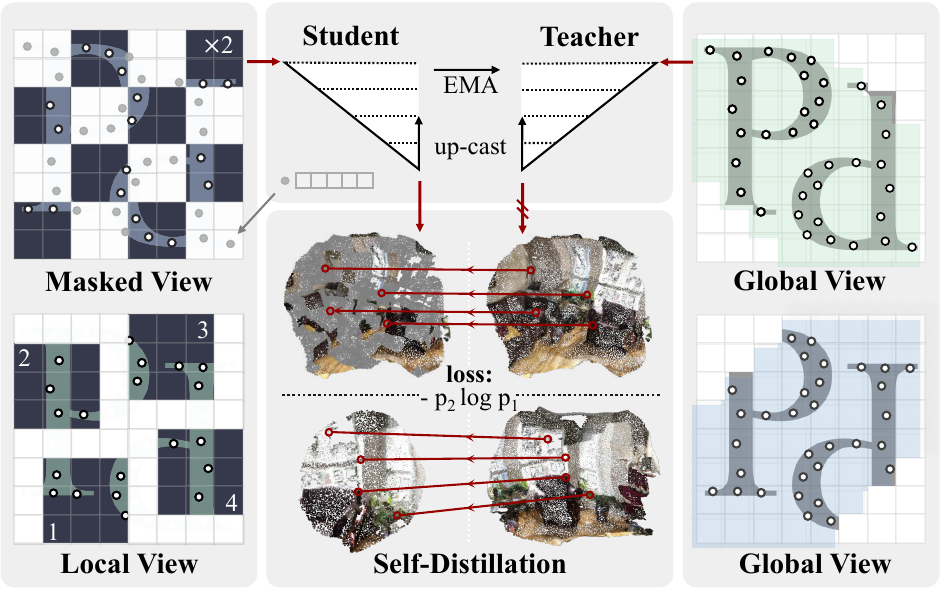}
    \vspace{-8mm}
\caption{\textbf{Self-distillation framework of Sonata.} (1) Local views (\textit{bottom left}) and global views (\textit{right}) are generated with dedicated spatial and photometric augmentations, while masked views are created by randomly masking out grid-based patches from the global views (\textit{top left}). (2) Embeddings from local and masked views are extracted by the student, with global views processed by the teacher (\textit{top}). (3) Points from local and masked views are matched with corresponding points in the global views based on their original spatial distance, allowing for the distillation of embeddings from global views to local and masked views (\textit{bottom}).}
    \label{fig:framework}
    \vspace{-6mm}
\end{figure}

However, the superior robustness of self-supervised representation is rooted in a core principle: \textit{continuously increasing the difficulty of pretext tasks as long as the model continues to converge}. This principle encourages enhancing the basic recipe with local-global view alignment and mask-unmask view alignment. Specifically, it involves aligning neighboring points from local views generated by small cropping ratios and masked views created with large masking ratios (see \figref{fig:framework} left), with unmasked global views that contain relatively richer information (see \figref{fig:framework} right). The difficulty of pretext tasks can be scaled by adjusting the cropping and masking ratios.

The more challenging the pretext tasks become, the higher the risk of model collapse. This instability calls for an \textit{asymmetric encoding approach}, specifically exponential moving average (EMA)~\cite{he2020moco}: rather than encoding all views with a shared-weight model, the approach encodes challenging local and masked views with an actively learned student model, while using a stable teacher model, updated with a moving average of the student parameters, to encode global views (see \figref{fig:framework} top). With the teacher preventing the student from being misled, the student is less likely to get lost in ``mission impossible" and has a greater chance to discover treasure within ``impossible" (\ie extreme 5\% crop ratio for local views and 70\% mask ratio for masked views).

In terms of SSL criteria, we move away from prior approaches that rely on contrastive and generative learning~\cite{wu2023msc}. Contrastive learning, limited by the number of point pairs in pairwise similarity computations, restricts scalability. Generative learning, meanwhile, partially anchors representations to predefined cues, limiting the model's capacity to capture more generalizable features. Following DINOv2, we adopt a self-distillation approach driven by Sinkhorn-Knopp centering~\cite{caron2020swav}, KoLeo regularization~\cite{sablayrolles2018spreading}, and clustering assignments~\cite{caron2020swav}. This adaptation initially exacerbates collapse into geometric shortcuts but holds greater potential for robust representation once this challenge is addressed.

\vspace{-2mm}
\subsection{Micro Design}
\label{sec:micro}
\vspace{-2mm}
We now discuss micro designs aimed at addressing geometric shortcuts. Since the problematic spatial information is inherently tied to point coordinates and directly used by operators, it is nearly impossible to mask out. This constraint defines the key strategies of our micro designs: \textit{obscuring spatial information} and \textit{emphasizing input features}.

\mypara{Decoder removal.} Previous approaches adhere to the original U-Net-style backbone for feature extraction; however, our observations in \secref{sec:pilot&principles} motivate us to remove the complex hierarchical decoder and perform self-distillation directly using the encoder’s output. This simple adjustment is key in our fight against geometric shortcuts, as it: (a) increases the feature channels participating in self-distillation (from 96 to 512), (b) streamlines the pipeline by involving fewer points in the pretext task after hierarchical pooling, and, most importantly, (c) introduces a natural way to obscure geometric cues: the positional information of points becomes naturally disturbed during hierarchical encoding and pooling. In fact, this removal proves to be even more beneficial than expected, boosting the linear probing result from 20.7\% to 60.4\%.

\mypara{Feature up-casting.} While the removal effectively prevents over-reliance on naive geometric cues, it also introduces certain limitations, particularly in leveraging multi-scale contexts. In the original U-Net structure, the decoder plays a key role in progressively aggregating features across scales to reconstruct semantic details. Without this process, self-distillation struggles to capture multi-scale spatial information and broader contextual relationships. To retain multi-scale features, we introduce a parameter-free feature up-casting process similar to hypercolumns in image segmentation~\cite{Hariharan_2015_CVPR}: progressively up-casting features back to the scale of the previous encoding stage, with the mapping relationships preserved through pooling layers and concatenation with features from the prior encoding stage. This approach provides richer, coarse-to-fine features from multi-scale encoding. While it does increase the risk of the model falling into geometric shortcuts, finding the right balance is key. Our ablation study shows that up-casting features twice achieve the best performance.

\mypara{Masked points jitter.} A slight random Gaussian jitter ($\sigma=0.005$) is applied to point coordinates as part of data augmentation. However, specifically for points to be masked, we additionally apply a stronger Gaussian jitter ($\sigma=0.01$) to further disrupt their spatial relationships. We pay special attention to these masked points because models are more likely to collapse to naive geometric cues, especially when point features are masked, making it difficult to derive solutions from neighboring unmasked points.

\mypara{Progressive parameter scheduler.} Geometric shortcuts are like traps on the convergence path of point cloud self-supervised learning. Similarly, we can set our own ``trap'' during training. For example, rather than starting with a challenging large mask size and mask ratio, we begin with a relatively small mask size (10 cm) and mask ratio (30\%), gradually increasing them to 40 cm ($0.1\Rightarrow 0.4$) and 70\% ($0.3\Rightarrow 0.7$) over the first 5\% of the training process. This strategy encourages the model to rely more on input features to solve the pre-defined pretext tasks, preventing it from shifting its reliance to point coordinates as training difficulty increases. This approach aligns with curriculum learning~\cite{bengio2009curriculum}, progressively challenging the model as it adapts. Similarly, in addition to the common learning rate scheduler, we also implement custom progressive schedulers for teacher temperature (0.04 $\Rightarrow$ 0.07) and weight decay (0.04 $\Rightarrow$ 0.2). We found that this design pushes these parameters to more extreme levels previously unexplored, further enhancing model performance.

\begin{table}[!t]
    \begin{minipage}{\columnwidth}
    \centering
        \tablestyle{4.8pt}{1.08}
        \input{table/data_collection}
        \vspace{-3mm}
        \caption{\textbf{Data source collection.}}\label{tab:data_collection}
        \vspace{1mm}
        \tablestyle{8pt}{1.08}
        \input{table/data_comparison}
        \vspace{-3mm}
        \caption{\textbf{Data scale comparison.}}\label{tab:data_comparison}
        \vspace{-5mm}
    \end{minipage}
\end{table}

\vspace{-2mm}
\subsection{Implementation and Evaluation Protocols}
\label{sec:implementation}
\vspace{-2mm}
In this section, we introduce the implementation details and the evaluation protocols for our experiments.

\mypara{Backbone.} We build our Sonata with Point Transformer V3 (PTv3)~\cite{wu2024ptv3} and refer to Pointcept~\cite{pointcept2023} for details of implementation. Building on this, we made an additional adjustment to enhance scalability: replacing all Batch Normalization (BN)~\cite{ioffe2015batch} layers with Layer Normalization (LN)~\cite{ba2016layer}. Although this replacement results in some initial accuracy degradation, it enhances domain adaptation by eliminating the need for additional domain-specific adjustments when scaling up with multi-dataset joint training~\cite{wu2024ppt}. Along with the scaling up of data, we also scale up the encoder block depths from \texttt{[2,2,2,6,2]} to \texttt{[3,3,3,12,3]} and widths from \texttt{[32,64,128,256,512]} to \texttt{[48,96, 192,384,512]}. This PTv3 model has 108M parameters.

\mypara{Data.} We extend the multi-dataset joint training approach introduced by PPT~\cite{wu2024ppt}, further expanding the data scale by removing the constraint of human labeling through unsupervised learning. This results in a collection of 140k scene-level point clouds from both real-world and simulated environments (outlined in \tabref{tab:data_collection}), making it 86.7$\times$ larger than the data scale of PointContrast~\cite{xie2020pointcontrast} and 5.9$\times$ larger than the data collection of PPT, as detailed in \tabref{tab:data_comparison}.

\mypara{Training.} We train Sonata on the 140k data collection for 200 epochs, using the AdamW optimizer~\cite{loshchilov2019adamw} with a batch size of 96, distributed across 32 GPUs. The learning rate linearly warms up over the first 10 epochs to a base value of 0.004, then decays following a cosine schedule~\cite{loshchilov2017sgdr}. Additionally, a layer-wise learning rate decay of 0.9 is applied to model parameters~\cite{zhou2022bert}. Weight decay is also controlled by a cosine schedule, progressively increasing from 0.04 to 0.2. For EMA, the student temperature is set to 0.1, while the teacher temperature gradually rises from 0.04 to 0.07 during the first 10 epochs~\cite{oquab2023dinov2}. The momentum starts at 0.994 and increases to 1 by the final iteration. For data augmentation and view generation, we follow the augmentation pipeline designed by MSC~\cite{wu2023msc}. We generate 2 global views (sampling 40\% to 100\% of scene points) and 4 local views (sampling 5\% to 40\% of scene points) for training, with 2 masked views generated based on the global views.

\mypara{Evaluation.} We evaluate the quality of the learned representation using the following three protocols after initializing the encoder with Sonata:
\begin{itemize}[leftmargin=5mm, itemsep=0mm, topsep=0mm, partopsep=0mm]
    \item In \textit{linear probing}, we keep the encoder frozen and up-cast the features to their original scale. A single linear layer, comprising less than 0.2\% of the total parameters, is then used to adapt these features to downstream tasks.
    \item In \textit{decoder probing}, we take a step back and reintroduce a lightweight hierarchical decoder, which accounts for 13\% of the total parameters. We then freeze the encoder, allowing only the decoder to actively learn.
    \item In \textit{full fine-tuning}, we follow the traditional approach by unfreezing the entire PTv3 U-Net-style backbone, tuning the learned representation to downstream tasks.
\end{itemize}
We advocate linear probing as the primary evaluation criterion for point SSL, considering other methods as intermediate steps. We look forward to the day when a linear-probed self-supervised model outperforms a fully fine-tuned one.

\begin{figure}[t]
    \centering
    \includegraphics[width=\linewidth]{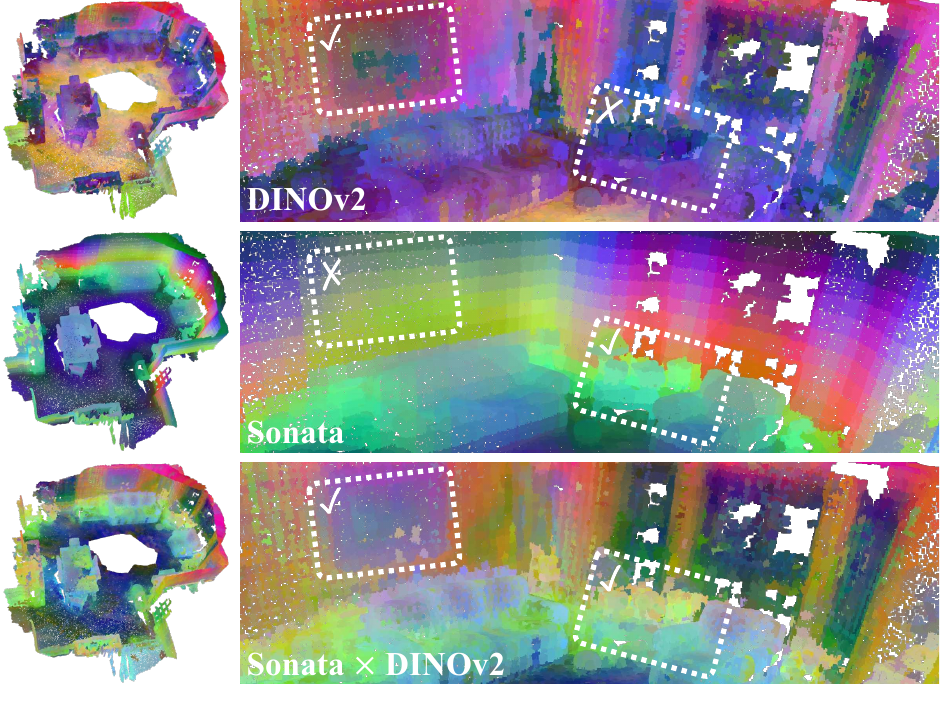}
    \vspace{-8mm}
\caption{\textbf{Zero-shot comparison with DINOv2.} We compare the PCA visualizations of DINOv2, Sonata, and their combined feature representation. DINOv2 excels at capturing photometric details, while Sonata better distinguishes spatial information. The combined model demonstrates improved coherence and detail, showcasing the complementary strengths of both models.}
    \label{fig:pca_2dx3d}
    \vspace{-4mm}
\end{figure}

%% file: table/data_collection.tex
\begin{tabular}{l|c|rrr|r}\toprule
Dataset &Source &Train &Val &Test &Total \\\midrule
ScanNet~\cite{dai2017scannet} &real &1,201 &312 &100 &1,613 \\
ScanNet++~\cite{yeshwanth2023scannet++} &real &712 &178 &126 &1,016 \\
S3DIS~\cite{armeni2016s3dis} &real &204 &68 &0 &272 \\
ArkitScenes~\cite{dehghan2021arkitscenes} &real &4,498 &549 &0 &5,047 \\
HM3D~\cite{ramakrishnan2021hm3d} &real &8,881 &1,119 &0 &10,000 \\
Structured3D~\cite{zheng2020structured3d} &sim. &18,348 &1,776 &1,697 &21,821 \\
ASE~\cite{avetisyan2024scenescript} &sim. &90,000 &10,000 &0 &100,000 \\ 
\cmidrule(lr){1-6}
\cellcolor[HTML]{efefef}Sonata~(ours) &\cellcolor[HTML]{efefef}mixed &\cellcolor[HTML]{efefef}123,844 &\cellcolor[HTML]{efefef}14,002 &\cellcolor[HTML]{efefef}1,923 &\cellcolor[HTML]{efefef}\textbf{139,769} \\
\bottomrule
\end{tabular}

%% file: table/data_comparison.tex
\begin{tabular}{l|rrr|l}\toprule
Method &Real &Sim &Total &Multipler \\\midrule
PC~\cite{xie2020pointcontrast} &1,613 &0 &1,613 &$\times$1 \\
MSC~\cite{wu2023msc} &6,660 &0 &6,660 &$\times$4.1 \\
PPT~\cite{wu2024ppt} &1,885 &21,821 &23,706 &$\times$14.7 \\
\cellcolor[HTML]{efefef}Sonata (ours) &\cellcolor[HTML]{efefef}17,948 &\cellcolor[HTML]{efefef}121,821 &\cellcolor[HTML]{efefef}\textbf{139,768} &\cellcolor[HTML]{efefef}\textbf{$\times$86.7} \\
\bottomrule
\end{tabular}

%% file: section/4_experiments.tex
\begin{table*}
    \begin{minipage}{0.44\textwidth}
    \centering
        \tablestyle{2.3pt}{1.08}
        \input{table/2dx3d}
        \vspace{-3mm}
        \caption{\textbf{Numerical comparison with DINO series.}}\label{tab:2dx3d}
        \vspace{1mm}
    \end{minipage}
    \hspace{2mm}
    \begin{minipage}{0.54\textwidth}
    \centering
        \tablestyle{3pt}{0.98
        }
        \input{table/data_efficiency}
        \vspace{-3mm}
        \caption{\textbf{Data efficiency.}}\label{tab:data_efficiency}
        \vspace{1mm}
    \end{minipage} \\
    \begin{minipage}{0.98\textwidth}
    \centering
        \tablestyle{2.6pt}{1.08}
        \input{table/parameter_efficiency}
        \vspace{-3mm}
        \caption{\textbf{Parameter efficiency.}}\label{tab:parameter_efficiency}
        \vspace{1mm}
    \end{minipage}
    \vspace{-6mm}
\end{table*}

\vspace{-2mm}
\section{Main Results}
\vspace{-2mm}
We validate the reliability of the Sonata representation using the evaluation protocols discussed in \secref{sec:implementation} and analyze the main properties based on these results.

\mypara{Comparison with image self-supervised model.} In \tabref{tab:2dx3d}, we compare the linear probing and decoder probing results on ScanNet and ScanNet200 semantic segmentation with the linear probing accuracy of representations transferred from the image self-supervised models DINOv2~\cite{oquab2023dinov2} and DINOv2.5~\cite{darcet2023vitneedreg}. Specifically, we aggregate unprojected pixel embeddings using ground truth camera poses and depth calculated by ray intersection with a reconstructed mesh, which provides more accuracy than sensor per-frame depth. Our results indicate that while the DINOs' representation demonstrates impressive robustness, Sonata offers a more suitable representation for 3D tasks, achieving +9.2\% on ScanNet and +1.5\% on ScanNet200 for semantic segmentation. Furthermore, combining Sonata with the DINOs yields higher accuracy than any single data modality alone (+3.9\% and +7.7\%, respectively), underscoring the promising potential of cross-modal self-distillation. Additional zero-shot comparisons (see \figref{fig:pca_2dx3d}) through PCA visualizations further corroborate these numerical findings.

\begin{table*}[!t]
    \begin{minipage}{0.98\textwidth}
    \centering
        \tablestyle{2.6pt}{1.1}
        \input{table/semantic_segmentation}
        \vspace{-3mm}
        \caption{\textbf{Indoor semantic segmentation.}}\label{tab:semantic_segmentation}
        \vspace{2mm}
    \end{minipage} \\
    \begin{minipage}{0.98\textwidth}
    \centering
        \tablestyle{2.6pt}{1.1}
        \input{table/instance_segmentation}
        \vspace{-3mm}
        \caption{\textbf{Indoor instance segmentation.}}\label{tab:instance_segmentation}
        \vspace{1mm}
    \end{minipage} \\
    \vspace{-6mm}
\end{table*}

\begin{table}[!ht]
    \begin{minipage}{0.48\textwidth}
    \centering
        \vspace{1mm}
        \tablestyle{0.3pt}{1.1}
        \input{table/outdoor_semantic_segmentation}
        \vspace{-3mm}
        \caption{\textbf{Outdoor semantic segmentation.}}\label{tab:outdoor_semantic_segmentation}
        \vspace{-6mm}
    \end{minipage}
\end{table}

\mypara{Data efficiency.} In \tabref{tab:data_efficiency}, we present the semantic segmentation performance of Sonata when probed or fine-tuned on the ScanNet dataset with limited scenes and annotations~\cite{hou2021csc}. The results demonstrate the exceptional data efficiency of Sonata, with improvements of 19.5\% in extreme data scarcity (1\% of scenes) and 10.4\% in limited annotation scenarios (20 points per scene) compared to training from scratch. Notably, even linear probing surpasses previous SOTA by a substantial margin (12.5\% with 1\% scenes), highlighting Sonata’s reliability in low-data scenarios.

\mypara{Parameter efficiency.} In \tabref{tab:parameter_efficiency}, we demonstrate parameter efficiency using both linear and decoder probing across various indoor semantic segmentation benchmarks, including ScanNet~\cite{dai2017scannet}, ScanNet200~\cite{rozenberszki2022scannet200}, ScanNet++~\cite{yeshwanth2023scannet++}, S3DIS Area5\cite{armeni2016s3dis}, and S3DIS 6-fold cross-validation~\cite{qi2017pointnet}. Semantic segmentation is emphasized as it provides a direct measure of point cloud representation quality. Results show that a single linear layer with a negligible number of parameters (\tinyless0.02\% of total parameters) is sufficient for Sonata to achieve strong performance on these benchmarks (\eg, 72.5\% on ScanNet and 73.4\% on S3DIS Area5). Furthermore, probing a decoder with only 13\% of the model’s parameters yields even higher accuracy (\eg, 79.1\% on ScanNet and 81.5\% on S3DIS 6-fold cross-validation). %
However, while decoder probing achieves SOTA results on ScanNet (20 classes) and S3DIS (13 classes), performance on ScanNet200 (200 classes) and ScanNet++ (100 classes) %
remains limited. This shows a limitation of the learned representation in distinguishing a large number of classes.

\mypara{Indoor semantic segmentation.} In \tabref{tab:semantic_segmentation}, we further enhance Sonata's semantic segmentation accuracy through full fine-tuning, consistently pushing SOTA results to new heights across the five widely recognized benchmarks, \eg, achieving 79.4\% on ScanNet and 82.3\% on S3DIS 6-fold cross-validation. However, we view full fine-tuning as an intermediate step toward a future where linear probing surpasses it. At present, full fine-tuning remains essential to achieve the highest performance on these benchmarks and close the remaining gap of 7.1\% and 5.8\% respectively.

\mypara{Indoor instance segmentation.} In \tabref{tab:instance_segmentation}, We also validate the robustness of Sonata representation on indoor instance segmentation benchmarks, including ScanNet~\cite{dai2017scannet}, ScanNet200~\cite{rozenberszki2022scannet200}, ScanNet++~\cite{yeshwanth2023scannet++}, and S3DIS~\cite{armeni2016s3dis}. Consistent with our findings in semantic segmentation, Sonata demonstrates strong parameter efficiency, achieving significant improvements with linear probing (10$\times$ mAP50 on ScanNet and 21$\times$ on ScanNet200) and decoder probing (12$\times$ mAP50 on ScanNet and 33$\times$ on ScanNet200). Full fine-tuning further boosts these results, achieving SOTA pre-training performance across benchmarks.
This clearly demonstrates that, unlike previous SSL approaches, the Sonata representation encodes instance-level information.

\mypara{Outdoor semantic segmentation.}
In \tabref{tab:outdoor_semantic_segmentation}, we adapt pre-training paradigm of Sonata to outdoor LiDAR scenarios through joint training on nuScenes~\cite{caesar2020nuscenes}, Waymo~\cite{sun2020waymo}, and SemanticKITTI~\cite{behley2019semantickitti} and evaluate semantic segmentation performance using our evaluation protocols. With linear probing, Sonata sets a robust parameter efficiency baseline. Decoder probing achieves significant gains. In full fine-tuning, Sonata surpasses the supervised PPT~\cite{wu2024ppt}, establishing new SOTA mIoU scores of 81.7, 72.9, and 72.6 across these benchmarks, underscoring the effectiveness of Sonata in outdoor perception tasks. Note that most of that performance can be recovered by more efficient decoder-only probing with a 95\%, 97\%, and 94\% of full fine-tuning.

\begin{figure*}[t]
    \centering
    \includegraphics[width=\linewidth]{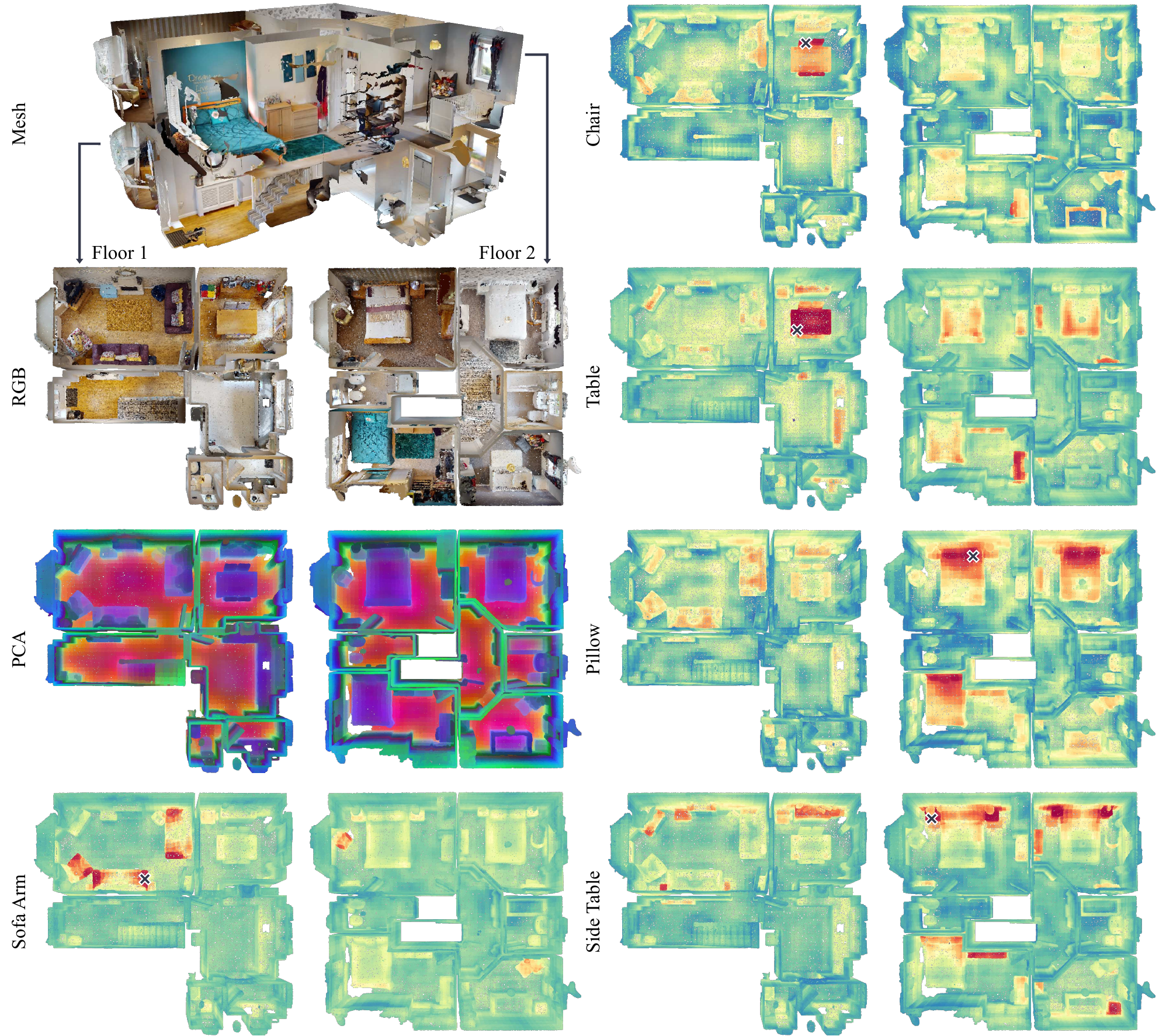}
    \vspace{-6mm}
    \caption{\textbf{Zero-shot representation across scenes.} We provide PCA-mapped colors and dense matching (with five representative points marked with \textcolor{darkblue}{{$\times$}}) on a house-scale point cloud from HM3D~\cite{ramakrishnan2021hm3d}, comprising 2 floors and 12 rooms (\textit{left:} floor 1, \textit{right:} floor 2). The visualization demonstrates that Sonata consistently delivers semantically rich and informative representations across diverse indoor environments.\looseness=-1}
    \label{fig:hm3d}
    \vspace{-4mm}
\end{figure*}

\mypara{Zero-shot representation across scenes.}
In \figref{fig:hm3d}, we visualize cross-scene zero-shot representation, including PCA and dense matching, using the Habitat-Matterport 3D Dataset (HM3D)~\cite{ramakrishnan2021hm3d}. Specifically, we utilize a house-scale point cloud spanning 2 floors and 12 rooms, showcasing a variety of indoor layouts and environments. Each room is separately encoded with Sonata, and the learned representations are visualized using PCA-mapped colors to highlight the semantic structure. Additionally, we select five representative points from various objects, including sofa arm, chair, table, pillow, and side table, and visualize dense matching by computing the similarity of each selected point with the rest of the house-scale point cloud. This process highlights the semantic coherence and clustering of features across objects and spaces. The visualization demonstrates that Sonata consistently provides semantically rich and informative representations across diverse indoor environments. These representations effectively capture distinct object patterns, exhibit a high degree of semantic granularity, and enable meaningful queries without any supervision, reinforcing the robustness and utility of the Sonata features.

%% file: table/2dx3d.tex
\begin{tabular}{lrrrrrrr}\toprule
2D $\times$ 3D &\multicolumn{3}{c}{ScanNet Val~\cite{dai2017scannet}} &\multicolumn{3}{c}{ScanNet200 Val~\cite{dai2017scannet}} \\
\cmidrule(lr){1-1} \cmidrule(lr){2-4} \cmidrule(lr){5-7}
Methods &mIoU &mAcc &allAcc &mIoU &mAcc &allAcc \\\midrule
\pretrain DINOv2~(lin.)~\cite{oquab2023dinov2} &63.09 &75.50 &82.42 &27.42 &37.59 &72.80 \\
\pretrain DINOv2.5~(lin.)~\cite{darcet2023vitneedreg} &63.36 &75.94 &82.30 &27.75 &39.23 &72.53 \\
\cellcolor[HTML]{f3f7fc}\pretrain Sonata (lin.) &72.52 &83.11 &89.74 &29.25 &41.61 &81.15 \\
\cellcolor[HTML]{f3f7fc}\ \ \plus DINOv2 (lin.) &\cellcolor[HTML]{ecf8f1}75.91 &\cellcolor[HTML]{ecf8f1}85.36 &\cellcolor[HTML]{ecf8f1}91.25 &\cellcolor[HTML]{ecf8f1}36.67 &\cellcolor[HTML]{ecf8f1}46.98 &\cellcolor[HTML]{def3e6}\textbf{82.85} \\
\cellcolor[HTML]{f3f7fc}\ \ \plus DINOv2.5 (lin.) &\cellcolor[HTML]{def3e6}\textbf{76.44} &\cellcolor[HTML]{def3e6}\textbf{85.68} &\cellcolor[HTML]{def3e6}\textbf{91.33} &\cellcolor[HTML]{def3e6}\textbf{36.96} &\cellcolor[HTML]{def3e6}\textbf{48.23} &\cellcolor[HTML]{ecf8f1}82.77 \\
\cmidrule(lr){1-7}
\cellcolor[HTML]{f3f7fc}\pretrain Sonata (dec.) &79.07 &86.57 &\cellcolor[HTML]{def3e6}\textbf{92.68} &33.54 &44.48 &\cellcolor[HTML]{def3e6}\textbf{84.07} \\
\cellcolor[HTML]{f3f7fc}\ \ \plus DINOv2 (dec.) &\cellcolor[HTML]{ecf8f1}79.12 &\cellcolor[HTML]{def3e6}\textbf{87.23} &92.47 &\cellcolor[HTML]{ecf8f1}37.73 &\cellcolor[HTML]{def3e6}\textbf{49.38} &83.31 \\
\cellcolor[HTML]{f3f7fc}\ \ \plus DINOv2.5 (dec.) &\cellcolor[HTML]{def3e6}\textbf{79.19} &\cellcolor[HTML]{ecf8f1}86.66 &\cellcolor[HTML]{ecf8f1}92.50 &\cellcolor[HTML]{def3e6}\textbf{38.27} &\cellcolor[HTML]{ecf8f1}48.57 &\cellcolor[HTML]{ecf8f1}83.77 \\
\bottomrule
\end{tabular}

%% file: table/data_efficiency.tex
\begin{tabular}{lrrrrrrrrrrr}\toprule
Data Efficiency &\multicolumn{5}{c}{Limited Scenes (Pct.)} &\multicolumn{5}{c}{Limited Annotation (Pts.)} \\
\cmidrule(lr){1-1} \cmidrule(lr){2-6} \cmidrule(lr){7-11}
Methods &1\% &5\% &10\% &20\% &Full &20 &50 &100 &200 &Full \\\midrule
\scratch SparseUNet~\cite{choy20194d} &26.0 &47.8 &56.7 &62.9 &\textcolor{darkgray}{72.2} &41.9 &53.9 &62.2 &65.5 &\textcolor{darkgray}{72.2} \\
\ \ \pretrain CSC~\cite{hou2021csc} &28.9 &49.8 &59.4 &64.6 &\textcolor{darkgray}{73.8} &55.5 &60.5 &65.9 &68.2 &\textcolor{darkgray}{73.8} \\
\ \ \pretrain MSC~\cite{wu2023msc} &29.2 &50.7 &61.0 &64.9 &\textcolor{darkgray}{75.4} &61.0 &65.6 &68.9 &69.6 &\textcolor{darkgray}{75.4} \\
\cmidrule(lr){1-11}
\scratch PTv2~\cite{wu2022ptv2} &24.8 &48.1 &59.8 &66.3 &\textcolor{darkgray}{75.4} &58.4 &66.1 &70.3 &71.2 &\textcolor{darkgray}{75.4} \\
\scratch PTv3~\cite{wu2024ptv3} &25.8 &48.9 &61.0 &67.0 &\textcolor{darkgray}{77.2} &60.1 &67.9 &71.4 &72.7 &\textcolor{darkgray}{77.2} \\
\ \ \pretrain PPT~\cite{wu2024ppt}~(sup.) &31.1 &52.6 &63.3 &68.2 &\textcolor{darkgray}{78.2} &62.4 &69.1 &\cellcolor[HTML]{fbfdfc}74.3 &\cellcolor[HTML]{fbfdfc}75.5 &\textcolor{darkgray}{78.2} \\
\cellcolor[HTML]{f3f7fc}\ \ \pretrain Sonata (lin.) &\cellcolor[HTML]{fbfdfc}43.6 &\cellcolor[HTML]{fbfdfc}62.5 &\cellcolor[HTML]{fbfdfc}68.6 &\cellcolor[HTML]{fbfdfc}69.8 &\textcolor{darkgray}{72.5} &\cellcolor[HTML]{fbfdfc}69.0 &\cellcolor[HTML]{fbfdfc}70.5 &71.1 &71.5 &\textcolor{darkgray}{72.5} \\
\cellcolor[HTML]{f3f7fc}\ \ \pretrain Sonata (dec.) &\cellcolor[HTML]{ecf8f1}44.5 &\cellcolor[HTML]{ecf8f1}64.1 &\cellcolor[HTML]{ecf8f1}69.8 &\cellcolor[HTML]{ecf8f1}72.5 &\textcolor{darkgray}{79.1} &\cellcolor[HTML]{ecf8f1}69.8 &\cellcolor[HTML]{ecf8f1}73.1 &\cellcolor[HTML]{ecf8f1}75.0 &\cellcolor[HTML]{ecf8f1}76.3 &\textcolor{darkgray}{79.1} \\
\cellcolor[HTML]{f3f7fc}\ \ \pretrain Sonata (full) &\cellcolor[HTML]{def3e6}\textbf{45.3} &\cellcolor[HTML]{def3e6}\textbf{65.7} &\cellcolor[HTML]{def3e6}\textbf{72.4} &\cellcolor[HTML]{def3e6}\textbf{72.8} &\textcolor{darkgray}{79.4} &\cellcolor[HTML]{def3e6}\textbf{70.5} &\cellcolor[HTML]{def3e6}\textbf{73.6} &\cellcolor[HTML]{def3e6}\textbf{76.0} &\cellcolor[HTML]{def3e6}\textbf{77.0} &\textcolor{darkgray}{79.4} \\
\bottomrule
\end{tabular}

%% file: table/parameter_efficiency.tex
\begin{tabular}{lrrccccccccccccccc}\toprule
Param. Effciency &\multicolumn{2}{c}{Params} &\multicolumn{3}{c}{ScanNet Val~\cite{dai2017scannet}} &\multicolumn{3}{c}{ScanNet200 Val~\cite{rozenberszki2022scannet200}} &\multicolumn{3}{c}{ScanNet++ Val~\cite{yeshwanth2023scannet++}} &\multicolumn{3}{c}{S3DIS Area 5~\cite{armeni2016s3dis}} &\multicolumn{3}{c}{S3DIS 6-fold~\cite{armeni2016s3dis}} \\
\cmidrule(lr){1-1} \cmidrule(lr){2-3} \cmidrule(lr){4-6} \cmidrule(lr){7-9} \cmidrule(lr){10-12} \cmidrule(lr){13-15} \cmidrule(lr){16-18}
Methods &\multicolumn{1}{c}{Learn.} &\multicolumn{1}{c}{Pct.} &mIoU &mAcc &allAcc &mIoU &mAcc &allAcc &mIoU &mAcc &allAcc &mIoU &mAcc &allAcc &mIoU &mAcc &allAcc \\\midrule
\scratch \textcolor{darkgray}{SparseUNet}~\cite{choy20194d} &\textcolor{darkgray}{39.2M} &\textcolor{darkgray}{100\%} &\textcolor{darkgray}{72.3} &\textcolor{darkgray}{80.2} &\textcolor{darkgray}{90.0} &\textcolor{darkgray}{25.0} &\textcolor{darkgray}{32.9} &\textcolor{darkgray}{80.4} &\textcolor{darkgray}{28.8} &\textcolor{darkgray}{38.4} &\textcolor{darkgray}{80.1} &\textcolor{darkgray}{66.3} &\textcolor{darkgray}{72.5} &\textcolor{darkgray}{89.8} &\textcolor{darkgray}{72.4} &\textcolor{darkgray}{80.9} &\textcolor{darkgray}{89.9} \\
\ \ \pretrain PC~\cite{xie2020pointcontrast} (lin.) &\tinyless0.2M &\tinyless0.1\% &5.6 &9.7 &50.0 &0.5 &0.9 &40.3 &1.8 &3.1 &46.4 &11.4 &18.6 &52.3 &11.7 &19.0 &51.2 \\
\ \ \pretrain CSC~\cite{hou2021csc} (lin.) &\tinyless0.2M &\tinyless0.1\% &12.6 &18.1 &64.2 &1.3 &2.1 &53.0 &2.8 &4.5 &53.6 &24.4 &32.0 &66.4 &24.9 &32.5 &66.9 \\
\ \ \pretrain MSC~\cite{wu2023msc} (lin.) &\tinyless0.2M &\tinyless0.1\% &14.1 &20.3 &62.9 &1.5 &2.5 &53.6 &4.5 &6.6 &61.3 &27.9 &35.5 &\cellcolor[HTML]{fbfdfc}71.1 &29.9 &37.9 &71.3 \\
\cmidrule(lr){1-18}
\scratch \textcolor{darkgray}{PTv3}~\cite{wu2024ptv3} &\textcolor{darkgray}{124.8M} &\textcolor{darkgray}{100\%} &\textcolor{darkgray}{77.6} &\textcolor{darkgray}{85.0} &\textcolor{darkgray}{92.0} &\textcolor{darkgray}{35.3} &\textcolor{darkgray}{46.0} &\textcolor{darkgray}{83.4} &\textcolor{darkgray}{42.1} &\textcolor{darkgray}{53.4} &\textcolor{darkgray}{85.6} &\textcolor{darkgray}{73.4} &\textcolor{darkgray}{78.9} &\textcolor{darkgray}{91.7} &\textcolor{darkgray}{77.7} &\textcolor{darkgray}{85.3} &\textcolor{darkgray}{91.5} \\
\ \ \pretrain MSC~\cite{wu2023msc} (lin.) &\tinyless0.2M &\tinyless0.2\% &\cellcolor[HTML]{fbfdfc}21.8 &\cellcolor[HTML]{fbfdfc}32.2 &\cellcolor[HTML]{fbfdfc}65.5 &\cellcolor[HTML]{fbfdfc}3.3 &\cellcolor[HTML]{fbfdfc}5.5 &\cellcolor[HTML]{fbfdfc}57.5 &\cellcolor[HTML]{fbfdfc}8.1 &\cellcolor[HTML]{fbfdfc}11.9 &\cellcolor[HTML]{fbfdfc}64.7 &\cellcolor[HTML]{fbfdfc}32.1 &\cellcolor[HTML]{fbfdfc}42.4 &70.9 &\cellcolor[HTML]{fbfdfc}34.6 &\cellcolor[HTML]{fbfdfc}46.0 &\cellcolor[HTML]{fbfdfc}71.3 \\
\cellcolor[HTML]{f3f7fc}\ \ \pretrain Sonata (lin.) &\tinyless0.2M &\tinyless0.2\% &\cellcolor[HTML]{ecf8f1}72.5 &\cellcolor[HTML]{ecf8f1}83.1 &\cellcolor[HTML]{ecf8f1}89.7 &\cellcolor[HTML]{ecf8f1}29.3 &\cellcolor[HTML]{ecf8f1}41.6 &\cellcolor[HTML]{ecf8f1}81.2 &\cellcolor[HTML]{ecf8f1}37.3 &\cellcolor[HTML]{ecf8f1}50.9 &\cellcolor[HTML]{ecf8f1}84.3 &\cellcolor[HTML]{ecf8f1}72.3 &\cellcolor[HTML]{ecf8f1}81.2 &\cellcolor[HTML]{ecf8f1}90.9 &\cellcolor[HTML]{ecf8f1}76.5 &\cellcolor[HTML]{ecf8f1}87.4 &\cellcolor[HTML]{ecf8f1}90.8 \\
\cellcolor[HTML]{f3f7fc}\ \ \pretrain Sonata (dec.) &16.3M &13\% &\cellcolor[HTML]{def3e6}\textbf{79.1} &\cellcolor[HTML]{def3e6}\textbf{86.6} &\cellcolor[HTML]{def3e6}\textbf{92.7} &\cellcolor[HTML]{def3e6}\textbf{33.5} &\cellcolor[HTML]{def3e6}\textbf{44.5} &\cellcolor[HTML]{def3e6}\textbf{84.1} &\cellcolor[HTML]{def3e6}\textbf{40.9} &\cellcolor[HTML]{def3e6}\textbf{52.6} &\cellcolor[HTML]{def3e6}\textbf{86.3} &\cellcolor[HTML]{def3e6}\textbf{74.5} &\cellcolor[HTML]{def3e6}\textbf{80.4} &\cellcolor[HTML]{def3e6}\textbf{92.6} &\cellcolor[HTML]{def3e6}\textbf{81.5} &\cellcolor[HTML]{def3e6}\textbf{88.8} &\cellcolor[HTML]{def3e6}\textbf{93.0} \\
\bottomrule
\end{tabular}

%% file: table/semantic_segmentation.tex
\begin{tabular}{y{21.5mm}rz{7.5mm}ccccccccccccccc}\toprule
Indoor Sem. Seg &\multicolumn{2}{c}{Params} &\multicolumn{3}{c}{ScanNet Val~\cite{dai2017scannet}} &\multicolumn{3}{c}{ScanNet200 Val~\cite{rozenberszki2022scannet200}} &\multicolumn{3}{c}{ScanNet++ Val~\cite{yeshwanth2023scannet++}} &\multicolumn{3}{c}{S3DIS Area 5~\cite{armeni2016s3dis}} &\multicolumn{3}{c}{S3DIS 6-fold~\cite{armeni2016s3dis}} \\
\cmidrule(lr){1-1} \cmidrule(lr){2-3} \cmidrule(lr){4-6} \cmidrule(lr){7-9} \cmidrule(lr){10-12} \cmidrule(lr){13-15} \cmidrule(lr){16-18}
Methods &\multicolumn{1}{c}{Learn.} &\multicolumn{1}{c}{Pct.} &mIoU &mAcc &allAcc &mIoU &mAcc &allAcc &mIoU &mAcc &allAcc &mIoU &mAcc &allAcc &mIoU &mAcc &allAcc \\\midrule
\scratch \textcolor{darkgray}{SparseUNet}~\cite{choy20194d} &\textcolor{darkgray}{39.2M} &\textcolor{darkgray}{100\%} &\textcolor{darkgray}{72.3} &\textcolor{darkgray}{80.2} &\textcolor{darkgray}{90.0} &\textcolor{darkgray}{25.0} &\textcolor{darkgray}{32.9} &\textcolor{darkgray}{80.4} &\textcolor{darkgray}{28.8} &\textcolor{darkgray}{38.4} &\textcolor{darkgray}{80.1} &\textcolor{darkgray}{66.3} &\textcolor{darkgray}{72.5} &\textcolor{darkgray}{89.8} &\textcolor{darkgray}{72.4} &\textcolor{darkgray}{80.9} &\textcolor{darkgray}{89.9} \\
\ \ \pretrain PC~\cite{xie2020pointcontrast} &39.2M &100\% &72.3 &80.9 &90.1 &26.2 &33.0 &79.9 &29.2 &39.7 &82.7 &68.1 &73.5 &90.0 &74.7 &83.3 &90.6 \\
\ \ \pretrain CSC~\cite{hou2021csc} &39.2M &100\% &72.8 &81.0 &90.7 &26.9 &33.7 &80.6 &32.5 &41.1 &83.7 &70.7 &76.4 &90.8 &75.5 &84.0 &90.9 \\
\ \ \pretrain MSC~\cite{wu2023msc} &39.2M &100\% &75.7 &83.4 &91.3 &32.0 &41.6 &82.3 &39.4 &49.6 &84.9 &70.7 &76.1 &91.0 &77.4 &85.3 &91.5 \\
\cmidrule(lr){1-18}
\scratch \textcolor{darkgray}{PTv3}~\cite{wu2024ptv3} &\textcolor{darkgray}{124.8M} &\textcolor{darkgray}{100\%} &\textcolor{darkgray}{77.6} &\textcolor{darkgray}{85.0} &\textcolor{darkgray}{92.0} &\textcolor{darkgray}{35.3} &\textcolor{darkgray}{46.0} &\textcolor{darkgray}{83.4} &\textcolor{darkgray}{42.1} &\textcolor{darkgray}{53.4} &\textcolor{darkgray}{85.6} &\textcolor{darkgray}{73.4} &\textcolor{darkgray}{78.9} &\textcolor{darkgray}{91.7} &\textcolor{darkgray}{77.7} &\textcolor{darkgray}{85.3} &\textcolor{darkgray}{91.5} \\
\ \ \pretrain MSC~\cite{wu2023msc} &124.8M &100\% &\cellcolor[HTML]{fbfdfc}78.2 &\cellcolor[HTML]{fbfdfc}85.3 &\cellcolor[HTML]{fbfdfc}92.2 &\cellcolor[HTML]{fbfdfc}33.4 &\cellcolor[HTML]{fbfdfc}43.7 &\cellcolor[HTML]{fbfdfc}83.4 &\cellcolor[HTML]{fbfdfc}42.4 &\cellcolor[HTML]{fbfdfc}53.6 &\cellcolor[HTML]{fbfdfc}85.9 &\cellcolor[HTML]{fbfdfc}69.9 &\cellcolor[HTML]{fbfdfc}74.9 &\cellcolor[HTML]{fbfdfc}91.2 &\cellcolor[HTML]{fbfdfc}77.4 &\cellcolor[HTML]{fbfdfc}84.7 &\cellcolor[HTML]{fbfdfc}91.7 \\
\ \ \pretrain PPT~\cite{wu2024ppt}~(sup.) &124.8M &100\% &\cellcolor[HTML]{ecf8f1}78.6 &\cellcolor[HTML]{ecf8f1}85.9 &\cellcolor[HTML]{ecf8f1}92.3 &\cellcolor[HTML]{ecf8f1}36.0 &\cellcolor[HTML]{ecf8f1}46.2 &\cellcolor[HTML]{ecf8f1}83.8 &\cellcolor[HTML]{ecf8f1}43.3 &\cellcolor[HTML]{ecf8f1}55.7 &\cellcolor[HTML]{ecf8f1}86.4 &\cellcolor[HTML]{ecf8f1}74.3 &\cellcolor[HTML]{ecf8f1}80.1 &\cellcolor[HTML]{ecf8f1}92.0 &\cellcolor[HTML]{ecf8f1}80.8 &\cellcolor[HTML]{ecf8f1}87.7 &\cellcolor[HTML]{ecf8f1}92.6 \\
\cellcolor[HTML]{f3f7fc}\ \ \pretrain Sonata &124.8M &100\% &\cellcolor[HTML]{def3e6}\textbf{79.4} &\cellcolor[HTML]{def3e6}\textbf{86.1} &\cellcolor[HTML]{def3e6}\textbf{92.5} &\cellcolor[HTML]{def3e6}\textbf{36.8} &\cellcolor[HTML]{def3e6}\textbf{46.5} &\cellcolor[HTML]{def3e6}\textbf{84.4} &\cellcolor[HTML]{def3e6}\textbf{43.7} &\cellcolor[HTML]{def3e6}\textbf{55.8} &\cellcolor[HTML]{def3e6}\textbf{86.6} &\cellcolor[HTML]{def3e6}\textbf{76.0} &\cellcolor[HTML]{def3e6}\textbf{81.6} &\cellcolor[HTML]{def3e6}\textbf{93.0} &\cellcolor[HTML]{def3e6}\textbf{82.3} &\cellcolor[HTML]{def3e6}\textbf{89.9} &\cellcolor[HTML]{def3e6}\textbf{93.3} \\
\bottomrule
\end{tabular}

%% file: table/instance_segmentation.tex
\begin{tabular}{y{21.5mm}rz{7.5mm}x{9.05mm}x{9.05mm}x{9.05mm}x{9.05mm}x{9.05mm}x{9.05mm}x{9.05mm}x{9.05mm}x{9.05mm}x{9.05mm}x{9.05mm}x{9.05mm}}\toprule
Indoor Ins. Seg &\multicolumn{2}{c}{Params} &\multicolumn{3}{c}{ScanNet Val~\cite{dai2017scannet}} &\multicolumn{3}{c}{ScanNet200 Val~\cite{rozenberszki2022scannet200}} &\multicolumn{3}{c}{ScanNet++ Val~\cite{yeshwanth2023scannet++}} &\multicolumn{3}{c}{S3DIS Area 5~\cite{armeni2016s3dis}} \\
\cmidrule(lr){1-1} \cmidrule(lr){2-3} \cmidrule(lr){4-6} \cmidrule(lr){7-9} \cmidrule(lr){10-12} \cmidrule(lr){13-15}
Methods &\multicolumn{1}{c}{Learn.} &\multicolumn{1}{c}{Pct.} &mAP25 & mAP50 &mAP &mAP25 & mAP50 &mAP &mAP25 & mAP50 &mAP &mAP25 & mAP50 &mAP \\\midrule
\scratch \textcolor{darkgray}{PointGroup}~\cite{jiang2020pointgroup} &\textcolor{darkgray}{124.8M} &\textcolor{darkgray}{100\%} &\textcolor{darkgray}{77.5} &\textcolor{darkgray}{61.7} &\textcolor{darkgray}{40.9} &\textcolor{darkgray}{40.1} &\textcolor{darkgray}{33.2} &\textcolor{darkgray}{23.1} &\textcolor{darkgray}{36.7} &\textcolor{darkgray}{30.7} &\textcolor{darkgray}{20.9} &\textcolor{darkgray}{55.7} &\textcolor{darkgray}{49.4} &\textcolor{darkgray}{37.8} \\
\ \ \pretrain MSC (lin.) &\tinyless0.2M &\tinyless0.2\% &13.3 &5.3 &2.3 &2.3 &1.0 &0.4 &4.8 &2.6 &1.3 &19.0 &13.0 &9.7 \\
\cellcolor[HTML]{f3f7fc}\ \ \pretrain Sonata (lin.) &\tinyless0.2M &\tinyless0.2\% &\cellcolor[HTML]{ecf8f1}72.6 &\cellcolor[HTML]{ecf8f1}53.9 &\cellcolor[HTML]{ecf8f1}30.7 &\cellcolor[HTML]{ecf8f1}30.9 &\cellcolor[HTML]{ecf8f1}21.3 &\cellcolor[HTML]{ecf8f1}10.9 &\cellcolor[HTML]{ecf8f1}31.6 &\cellcolor[HTML]{ecf8f1}22.4 &\cellcolor[HTML]{ecf8f1}12.2 &\cellcolor[HTML]{ecf8f1}45.8 &\cellcolor[HTML]{ecf8f1}36.6 &\cellcolor[HTML]{ecf8f1}26.1 \\
\cellcolor[HTML]{f3f7fc}\ \ \pretrain Sonata (dec.) &16.3M &13\% &\cellcolor[HTML]{def3e6}\textbf{76.8} &\cellcolor[HTML]{def3e6}\textbf{62.8} &\cellcolor[HTML]{def3e6}\textbf{40.8} &\cellcolor[HTML]{def3e6}\textbf{40.8} &\cellcolor[HTML]{def3e6}\textbf{33.3} &\cellcolor[HTML]{def3e6}\textbf{22.8} &\cellcolor[HTML]{def3e6}\textbf{38.1} &\cellcolor[HTML]{def3e6}\textbf{29.1} &\cellcolor[HTML]{def3e6}\textbf{18.8} &\cellcolor[HTML]{def3e6}\textbf{63.7} &\cellcolor[HTML]{def3e6}\textbf{57.1} &\cellcolor[HTML]{def3e6}\textbf{45.1} \\
\cmidrule(lr){1-15}
\ \ \pretrain MSC (full) &124.8M &100\% &78.4 &62.9 &41.1 &40.5 &33.8 &23.4 &38.9 &30.9 &21.7 &56.3 &50.5 &38.1 \\
\ \ \pretrain PPT~\cite{wu2024ppt}~(sup.) &124.8M &100\% &\cellcolor[HTML]{ecf8f1}78.9 &\cellcolor[HTML]{ecf8f1}63.5 &\cellcolor[HTML]{ecf8f1}42.1 &\cellcolor[HTML]{ecf8f1}40.8 &\cellcolor[HTML]{ecf8f1}34.1 &\cellcolor[HTML]{ecf8f1}24.0 &\cellcolor[HTML]{ecf8f1}39.3 &\cellcolor[HTML]{ecf8f1}32.8 &\cellcolor[HTML]{ecf8f1}21.9 &\cellcolor[HTML]{ecf8f1}57.5 &\cellcolor[HTML]{ecf8f1}51.2 &\cellcolor[HTML]{ecf8f1}39.7 \\
\cellcolor[HTML]{f3f7fc}\ \ \pretrain Sonata (full) &124.8M &100\% &\cellcolor[HTML]{def3e6}\textbf{79.2} &\cellcolor[HTML]{def3e6}\textbf{63.9} &\cellcolor[HTML]{def3e6}\textbf{42.4} &\cellcolor[HTML]{def3e6}\textbf{42.1} &\cellcolor[HTML]{def3e6}\textbf{35.6} &\cellcolor[HTML]{def3e6}\textbf{25.4} &\cellcolor[HTML]{def3e6}\textbf{40.3} &\cellcolor[HTML]{def3e6}\textbf{33.6} &\cellcolor[HTML]{def3e6}\textbf{22.3} &\cellcolor[HTML]{def3e6}\textbf{63.8} &\cellcolor[HTML]{def3e6}\textbf{57.4} &\cellcolor[HTML]{def3e6}\textbf{45.5} \\
\bottomrule
\end{tabular}

%% file: table/outdoor_semantic_segmentation.tex
\resizebox{\linewidth}{!}{%
\begin{tabular}{lrrrrrrrrr}\toprule
Outdoor Sem. Seg. &\multicolumn{3}{c}{nuScenes Val~\cite{caesar2020nuscenes}} &\multicolumn{3}{c}{Waymo Val~\cite{sun2020waymo}} &\multicolumn{3}{c}{Sem.KITTI Val~\cite{behley2019semantickitti}} \\
\cmidrule(lr){1-1} \cmidrule(lr){2-4} \cmidrule(lr){5-7} \cmidrule(lr){8-10}
Methods &mIoU &mAcc &allAcc &mIoU &mAcc &allAcc &mIoU &mAcc &allAcc \\\midrule
\scratch \textcolor{darkgray}{PTv3}~\cite{wu2024ptv3} &\textcolor{darkgray}{80.4} &\textcolor{darkgray}{87.2} &\textcolor{darkgray}{94.7} &\textcolor{darkgray}{71.3} &\textcolor{darkgray}{80.5} &\textcolor{darkgray}{94.7} &\textcolor{darkgray}{69.1} &\textcolor{darkgray}{76.1} &\textcolor{darkgray}{92.6} \\
\cellcolor[HTML]{f3f7fc}\ \ \pretrain Sonata (lin.) &66.1 &77.2 &92.4 &60.5 &72.5 &92.5 &62.0 &72.5 &91.0 \\
\cellcolor[HTML]{f3f7fc}\ \ \pretrain Sonata (dec.) &\cellcolor[HTML]{def3e6}\textbf{77.3} &\cellcolor[HTML]{def3e6}\textbf{85.9} &\cellcolor[HTML]{def3e6}\textbf{94.2} &\cellcolor[HTML]{def3e6}\textbf{70.8} &\cellcolor[HTML]{def3e6}\textbf{78.8} &\cellcolor[HTML]{def3e6}\textbf{94.3} &\cellcolor[HTML]{def3e6}\textbf{68.4} &\cellcolor[HTML]{def3e6}\textbf{76.5} &\cellcolor[HTML]{def3e6}\textbf{92.3} \\
\cmidrule(lr){1-10}
\ \ \pretrain PPT~\cite{wu2024ppt}~(sup.) &81.2 &87.5 &94.8 &72.1 &81.3 &94.8 &72.3 &77.5 &93.4 \\
\cellcolor[HTML]{f3f7fc}\ \ \pretrain Sonata (full) &\cellcolor[HTML]{def3e6}\textbf{81.7} &\cellcolor[HTML]{def3e6}\textbf{87.9} &\cellcolor[HTML]{def3e6}\textbf{95.0} &\cellcolor[HTML]{def3e6}\textbf{72.9} &\cellcolor[HTML]{def3e6}\textbf{81.9} &\cellcolor[HTML]{def3e6}\textbf{94.9} &\cellcolor[HTML]{def3e6}\textbf{72.6} &\cellcolor[HTML]{def3e6}\textbf{77.9} &\cellcolor[HTML]{def3e6}\textbf{93.4} \\
\bottomrule
\end{tabular}
}

%% file: section/5_conclusion.tex
\vspace{-2mm}
\section{Conclusion and Discussion}
\vspace{-2mm} 

In this work, we make progress towards self-supervised learning of a strong and reliable 3D point representation, that can zero-shot correspond semantically similar 3D points, down to the instance level, even under the presence of spatial and visual perturbations. 
We demonstrate that such a representation can serve as the foundation for 3D tasks in semantic and instance-level grouping.
We find existing 3D self-supervised learning approaches lacking, and hypothesize that this is due to what we term the geometric shortcut, a problem unique to 3D, which causes representations to collapse to low-level spatial features. We demonstrate the deficiency of these collapsed representations through linear probing.
Beginning with a point self-distillation framework, we tackle the geometric shortcut by attaching SSL losses at coarser spatial scales, disturbing the spatial information of masked points with no features, and progressively increasing task difficulty to prevent over-reliance on accessible geometric cues. This change enables effective scaling up, ultimately composing a \textbf{Sonata} from 140k point clouds.
Sonata demonstrates semantically meaningful zero-shot visualization, as well as exceptional parameter and data efficiency. Full fine-tuning further advances SOTA across 3D indoor and outdoor perception tasks.

We discuss \textit{limitations and future works} as follows:
\begin{itemize}[leftmargin=5mm, itemsep=0mm, topsep=0mm, partopsep=0mm]
    \item \textit{Enhancing semantic meaning}. We believe there is significant potential to enhance the semantic richness of Sonata’s representations. Currently, our training does not yet leverage the vast resource of 1M object-level assets~\cite{liu2023openshape}, which could provide valuable augmentation for our scene-level point cloud dataset. Integrating these object-level point clouds could deepen the model’s semantic understanding by introducing finer object-specific details, creating a more robust foundation for scene-level and cross-instance semantics.
    \item \textit{Unifying training scenarios}. Unifying training across indoor and outdoor scenarios is a promising direction for future work. Currently, Sonata separates pre-training for each setting to focus on a reliable SSL framework without introducing the additional challenge of a domain gap. However, unification is feasible. The main challenges lie in point density and input features: point density can be aligned by scaling, while enhancing outdoor LiDAR data with color from lifted images and pseudo normal vectors based on LiDAR viewing direction helps bridge feature gaps. Additionally, applying randomized noise and masking on these features could further enhance generalization.
    \item \textit{Scaling with video data}. Natural 3D point cloud datasets have inherent scale limitations compared to video data. To address this, we aim to leverage video datasets in two ways: 1. using metric~\cite{bochkovskii2024depth} or stereo~\cite{wang2024dust3r} depth estimation to lift videos of static scenes into pixel-aligned point clouds, and 2. generating sparse point clouds from dynamic egocentric videos using SLAM algorithms~\cite{engel2023aria}. This approach opens new possibilities for training on large-scale, real-world diverse scenes.
    \item \textit{Cross-modal distillation}. Our evidence shows that self-supervised models from different modalities, like Sonata for point clouds and DINOv2~\cite{oquab2023dinov2} for images, capture complementary representations, and combining them leads to stronger representation. This suggests promising potential for cross-modal self-distillation to enhance both 3D and image representations. A straightforward approach would be to lift DINOv2 features into 3D and integrate them within Sonata’s pre-training paradigm. Additionally, developing a unified SSL framework with simultaneous self- and cross-modal distillation across point clouds and images could further enrich multi-modal representation learning.
\end{itemize}
We hope our insights with Sonata inspire innovations in reliable point self-supervised learning and pave the way for future research in 3D representations and their applications by reducing the reliance on extensive data and computational resources through reliable point representations.

\begin{figure}[t]
    \vspace{-0mm}
    \centering
    \includegraphics[width=\linewidth]{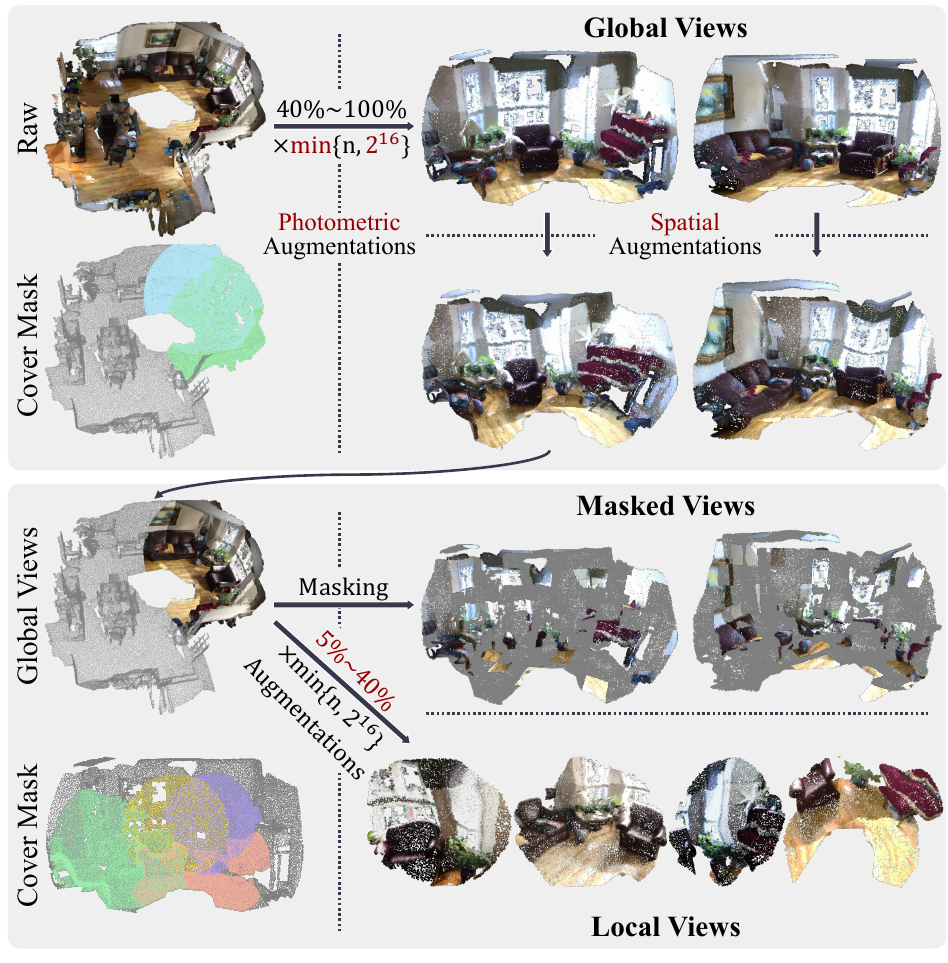}
    \vspace{-7.5mm}
    \caption{\textbf{View generation.} \textit{Top}: we generate global crops using random crops with a crop ratio ranging from 40\% to 100\% of the minimal number of the raw point cloud size and $2^{16}$, combined with random photometric and spatial augmentations. Photometric augmentation is shared among all global views, while spatial augmentation is applied independently to each global view to balance the challenges posed by spatial and photometric features. The first global view is designated as the principal view, and the center of the subsequent global view is restricted to fall within the principal view. \textit{Bottom:} Local views are generated with a similar pipeline as global views but with a crop ratio restricted to 5\% to 40\%. All augmentations are applied independently to each local view. Additionally, masked views are obtained by applying random patch masks to the global views.}
    \label{fig:view_generation}
    \vspace{-6mm}
\end{figure}

%% file: section/6_appendix.tex
For a thorough understanding of our Sonata, we have compiled a detailed Appendix. The table of contents below offers an overview and guide to specific sections of interest.

\hypersetup{linkbordercolor=black,linkcolor=black}

\setlength{\cftbeforesecskip}{0.5em}
\cftsetindents{section}{0em}{1.8em}
\cftsetindents{subsection}{1em}{2.5em}
\cftsetindents{subsubsection}{2em}{3.2em}

\etoctoccontentsline{part}{Appendix}
\localtableofcontents
\hypersetup{linkbordercolor=red,linkcolor=red}

\begin{table}[t]
\begin{algorithm}[H]
   \caption{point self-distillation pseudocode.}
   \label{algo:self_distillation}
    \definecolor{codeblue}{rgb}{0.25,0.5,0.5}
    \lstset{
      basicstyle=\fontsize{7.2pt}{7.2pt}\ttfamily\bfseries,
      commentstyle=\fontsize{7.2pt}{7.2pt}\color{codeblue},
      keywordstyle=\fontsize{7.2pt}{7.2pt},
    }
\begin{lstlisting}[language=python]
'''
To simplify, we present the pseudocode using a single
local(masked)-global pair.
'''

# gs, gt: student and teacher networks
# cs, ct: student and teacher online clustering head
# tps, tpt, student and teacher temperatures
# m: network momentum rates
# k: upcast level

# initialize student and teacher network and head
gt.params, ct.params = gs.params, cs.params
gt.requires_grad = False
ct.requires_grad = False

for p in loader:  # load a batch of point cloud
    # ps: local(mask) view, pt: global view
    ps, pt = view_generator(p)
    if ps is MaskedView:
        # apply gaussian noise to masked points
        ps.coord[p1.mask] += gaussian(s)
    # encode network feature
    fs, ft = gs(ps), gt(pt)
    # up-cast network feature
    fs, ft = upcast(fs, k), upcast(ft, k)
    # compute similarity with online cluster (SwAV)
    ss, st = cs(s1), ct(s2)
    # center with sinkhorn-knopp
    st = centering(st)

    # match neighbor point pairs with the original 
    # coordinate before augmentation, return index
    is, it = match(ps.origin_coord, pt.origin_coord)
    loss = H(ss[is], st[it])
    loss.backward()

    # update student and teacher network and head
    update(gs.params)
    update(cs.params)
    gt.params =  m*gt.params + (1-m)*gs.params
    ct.params =  m*ct.params + (1-m)*cs.params

def H(t, s):
    s = softmax(s / tps, dim=-1)
    # center with sinkhorn-knopp and sharpen
    t = softmax(center(t) / tpt, dim=-1)
    return - (t * log(s)).sum(dim=1).mean()
\end{lstlisting}
\end{algorithm}
\vspace{-10mm}
\end{table}
\vspace{1mm}
\section{Additional Implementation}
\subsection{View Generation} 
In \figref{fig:view_generation}, we illustrate the view generation pipeline of Sonata. Specifically, global views are generated with a random crop ratio between 40\% and 100\%, while local views use a ratio between 5\% and 40\%. The crop ratio is applied to the smaller of the raw point cloud size or $2^{16}$ points. The first global view is designated as the principal view, and subsequent global and local view centers are restricted to lie within this principal view. Random photometric and spatial augmentations~\cite{wu2023msc} are applied to all views. For global views, photometric augmentations are shared after being randomized, whereas spatial augmentations are applied independently to each view. Masked views are generated by applying random patch masks to the global views.

\vspace{-1mm}
\subsection{Point Self-distillation}
\begin{figure}[t]
    \centering
    \includegraphics[width=\linewidth]{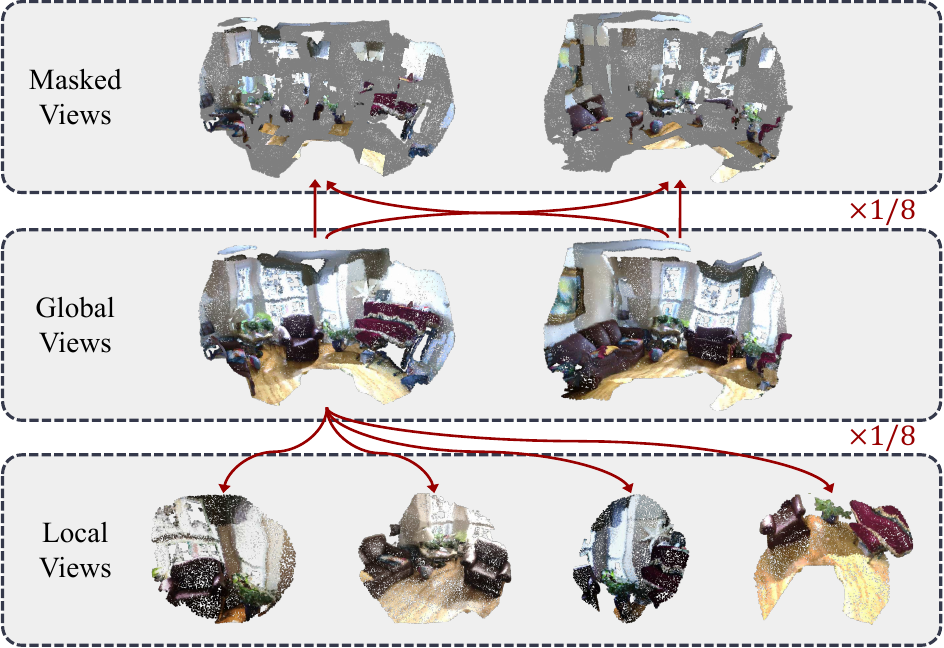}
    \vspace{-7mm}
    \caption{\textbf{Point self-distillation loss items.} The pair-wise point self-distillation between masked views and global views, and between local views and the principal global view. We evenly weight the loss terms for the 8 point self-distillation pairs.}
    \label{fig:point_distillation}
    \vspace{-4mm}
\end{figure}

\vspace{-1mm}
In Algo.\ref{algo:self_distillation}, we provide a simplified pseudocode for point self-distillation using a single local (masked)-global pair of random views. Additionally as visualized in \figref{fig:point_distillation}), we actually use a total of 4 local views, 2 masked views, and 2 global views. For the local views, point self-distillation is conducted between each local view and the principal global view. For the masked views, pair-wise point self-distillation is performed with each global view. These loss terms for all point self-distillation pairs are evenly weighted. \looseness=-1

\begin{table}[h]
    \begin{minipage}{0.48\textwidth}
    \centering
        \vspace{1mm}
        \tablestyle{6.5pt}{1.08}
        \input{table/aeo}
        \vspace{-3mm}
        \caption{\textbf{Out-of-distribution perception capability.} We evaluate this capability on the AEO dataset~\cite{straub2024efm3d} with 22 sparse SLAM point clouds, supervised by semantic labels from object bounding boxes.}\label{tab:ood}
        \vspace{-5mm}
    \end{minipage}
\end{table}

\section{Additional Properties}
\subsection{Out-of-distribution (OOD) Perception.}
\begin{figure}[t]
    \centering
    \includegraphics[width=\linewidth]{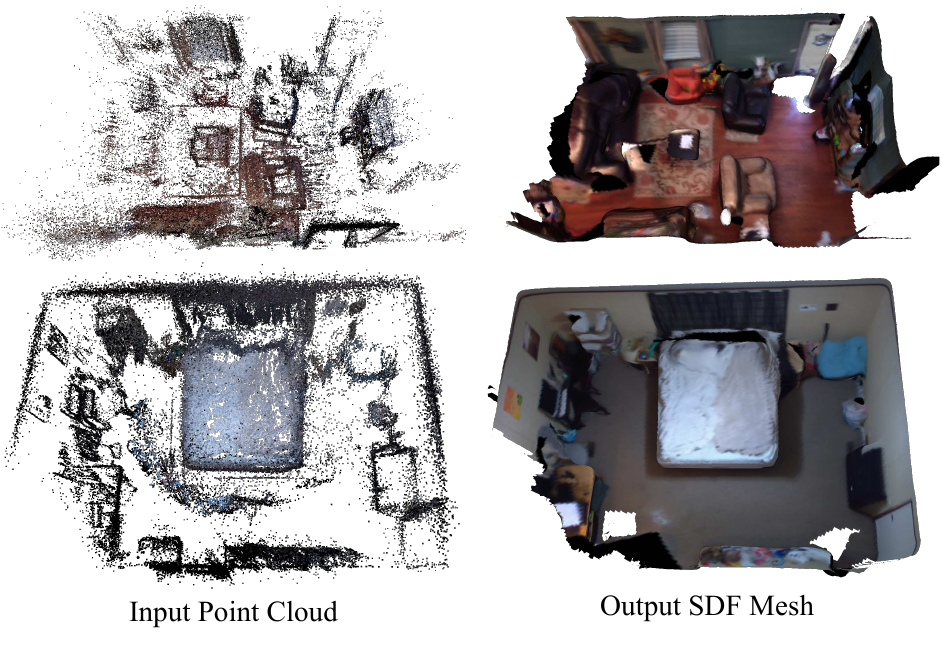}
    \vspace{-8mm}
    \caption{\textbf{Surface reconstruction.} Scene surface is reconstructed with SDF regression from frozen Sonata features, demonstrating strong geometric priors and cross-domain generalization.}
    \label{fig:surface_reconstruction}
    \vspace{-4mm}
\end{figure}

In \tabref{tab:ood}, we evaluate the out-of-distribution perception capability of Sonata using the Aria Everyday Objects (AEO) dataset~\cite{straub2024efm3d} for semantic segmentation. This dataset consists of 25 sparse SLAM-generated point clouds, each annotated with 17 object categories. Among these, three samples (IDs: 0, 5, 24) are reserved for validation, while the remaining 22 are used for training. This experimental setup presents significant challenges, including unseen data patterns (as shown by the sparse pattern of SLAM-generated point clouds in \figref{fig:surface_reconstruction}, left column), limited training data, and imprecise annotations derived from bounding box labels. We assess Sonata by performing both probing and fine-tuning on this semantic segmentation task, supervised by the transferred semantic labels.

The results show that the linear probing of Sonata achieves a mIoU of 32.0\%, which still has a gap of 2.9\% compared to the 34.9\% mIoU achieved by training from scratch. This indicates a current limitation of Sonata: insufficient diversity in training data patterns. Currently, we only include dense indoor point clouds, focusing on building a reliable point SSL framework without introducing additional domain gap challenges. However, fine-tuning Sonata demonstrates its robustness, achieving a remarkable 21.0\% improvement over training from scratch. This substantial leap underscores the strength and adaptability of Sonata representations, particularly in challenging OOD perception tasks. These findings further reinforce Sonata’s potential as a foundation for tackling tasks with limited or noisy training data in diverse and complex environments.

\subsection{Surface Reconstruction.}
Previous experiments have already demonstrated the rich semantic information encoded in Sonata representations. To further investigate whether Sonata also captures dense geometric priors, we conducted a surface regression experiment using frozen Sonata features. A Truncated Signed Distance Function (TSDF)~\cite{curless1996volumetric} volume is defined within a $4m \times 4m \times 4m$ local coordinate system with a resolution of $96\times96\times96$ ($\approx$4cm per voxel). The volume is patchified into $8\times8\times8$ patches, with each patch projected to a feature dimension of 512. This results in $512\times12\times12\times12$ volume tokens. We applied learned positional encodings to the patches, and tri-linear interpolation ensured consistency between the positional encodings and Sonata features. To decode the volume tokens into a dense TSDF volume, we simply use the standard transformer decoder~\cite{vaswani2017attention} implemented in PyTorch~\cite{paszke2019pytorch}, with the Sonata features being the ``memory'' and the voxel tokens being the decoding ``target''. One can also see the Sonata features as the context while the voxel tokens are the queries. After decoding, the outputs were reshaped to reconstruct the dense TSDF volume. This approach is inspired by the Large Reconstruction Models (LRM)s~\cite{hong2023lrm}. We employed the EVL training and TSDF fusion~\cite{straub2024efm3d, newcombe2011kinectfusion} pipeline, training the network on the synthetic ASE dataset~\cite{avetisyan2024scenescript}. The cross-domain generalization was tested on the SLAM-generated point cloud inputs of the AEO dataset, as illustrated in the \figref{fig:surface_reconstruction}. The results qualitatively demonstrate that dense scene geometry can be reconstructed solely from frozen Sonata features, showcasing learned geometric priors within Sonata representations.

\begin{table}[!t]
    \begin{minipage}{0.48\textwidth}
    \centering
        \tablestyle{10pt}{1.08}
        \input{table/appendix_scannet}
        \vspace{-3mm}
        \caption{\textbf{ScanNet V2 semantic segmentation.} }\label{tab:appendix_scannet}
        \vspace{-6mm}
    \end{minipage}
\end{table}

\begin{table}[!t]
    \begin{minipage}{0.48\textwidth}
    \centering
        \tablestyle{10pt}{1.08}
        \input{table/appendix_s3dis}
        \vspace{-3mm}
        \caption{\textbf{S3DIS semantic segmentation.} }\label{tab:appendix_s3dis}
        \vspace{-8mm}
    \end{minipage}
\end{table}

\section{Additional Comparision}
In this section, we expand the combined results table for indoor semantic segmentation from the main paper, providing a more detailed comparison of results on two key benchmarks: ScanNet~\cite{dai2017scannet} (see \tabref{tab:appendix_scannet}) and S3DIS~\cite{armeni2016s3dis} (see \tabref{tab:appendix_s3dis}). Specifically, the ScanNet v2 dataset contains 1,513 room scans reconstructed from RGB-D frames, with 1,201 scenes allocated for training and 312 for validation. The input point clouds are derived from the vertices of reconstructed meshes, where each point is labeled with one of 20 semantic categories (e.g., wall, floor, table). The S3DIS dataset includes 271 rooms distributed across six areas from three buildings, specifically designed for semantic scene parsing. Following established practices~\cite{tchapmi2017segcloud,qi2017pointnet++,zhao2021point}, area 5 is reserved for testing, and 6-fold cross-validation is performed across the remaining areas. Unlike ScanNet v2, S3DIS features densely sampled points on mesh surfaces, with annotations across 13 categories. For both datasets, we adopt the mean class-wise intersection over union (mIoU) as the primary metric to evaluate performance on indoor semantic segmentation tasks, adhering to standard conventions~\cite{qi2017pointnet++}. These expanded tables provide a detailed breakdown of performance metrics alongside the publication years of previous works, allowing readers to trace the evolution of advancements in 3D representation learning. Entries labeled as \scratch correspond to models trained from scratch, while \pretrain denotes results achieved using pre-trained models.

%% file: table/aeo.tex
\begin{tabular}{lrrrrrr}\toprule
Sem. Seg. &\multicolumn{2}{c}{Params} &\multicolumn{3}{c}{AEO~\cite{straub2024efm3d}} \\
\cmidrule(lr){1-1} \cmidrule(lr){2-3} \cmidrule(lr){4-6}
Methods &Learn. &Pct. &mIoU &mAcc &allAcc \\\midrule
\scratch PTv3 &124.8M &100\% & 34.91& 47.92& 63.79\\
\cellcolor[HTML]{f3f7fc}\ \ \pretrain Sonata (lin.) &\tinyless0.2M &\tinyless0.2\% & 32.03& 47.45& 53.25\\
\cellcolor[HTML]{f3f7fc}\ \ \pretrain Sonata (full) &124.8M &100\% &\cellcolor[HTML]{def3e6}\textbf{55.90} &\cellcolor[HTML]{def3e6}\textbf{63.49} &\cellcolor[HTML]{def3e6}\textbf{84.50} \\
\bottomrule
\end{tabular}

%% file: table/appendix_scannet.tex
\begin{tabular}{lccc}\toprule
Methods &Year &Val &Test \\\midrule
\scratch PointNet++~\cite{qi2017pointnet++} & 2017 & 53.5 & 55.7 \\
\scratch 3DMV~\cite{dai20183dmv} & 2018 & -& 48.4\\
\scratch PointCNN~\cite{li2018pointcnn} & 2018 & -& 45.8\\
\scratch SparseConvNet~\cite{graham20183d} & 2018 & 69.3 & 72.5 \\
\scratch PanopticFusion~\cite{narita2019panopticfusion} & 2019 & -& 52.9\\

\scratch PointConv~\cite{wu2019pointconv} & 2019 & 61.0 & 66.6\\
\scratch JointPointBased~\cite{chiang2019unified}& 2019 & 69.2 & 63.4\\
\scratch KPConv~\cite{thomas2019kpconv} & 2019 & 69.2 & 68.6 \\
\scratch PointASNL~\cite{yan2020pointasnl}& 2020 & 63.5 & 66.6\\
\scratch SegGCN~\cite{lei2020seggcn}& 2020 & -& 58.9\\
\scratch RandLA-Net~\cite{hu2020randla} & 2020 & - & 64.5 \\
\scratch JSENet~\cite{hu2020jsenet} & 2020 & - & 69.9 \\
\scratch FusionNet~\cite{zhang2020deep} & 2020 & - & 68.8 \\
\scratch FastPointTransformer~\cite{park2022fast} &2022 &72.4 &- \\
\scratch SratifiedTranformer~\cite{lai2022stratified} &2022 &74.3 &73.7 \\
\scratch PointNeXt~\cite{qian2022pointnext} &2022 &71.5 &71.2 \\
\scratch LargeKernel3D~\cite{chen2023largekernel3d} & 2023 &73.5 &73.9 \\
\scratch PointMetaBase~\cite{lin2023pointmetabase} & 2023 &72.8 &71.4 \\
\scratch PointConvFormer~\cite{wu2023pointconvformer} & 2023 &74.5 &74.9 \\
\scratch OctFormer~\cite{Wang2023OctFormer} &2023 &75.7 &76.6 \\
\scratch Swin3D~\cite{yang2023swin3d} &2023 &77.5 &77.9 \\
\ \ \pretrain Supervised~\cite{yang2023swin3d} &2023 &76.7 &77.9 \\
\scratch KPConvX~\cite{thomas2024kpconvx} &2024 &76.3 &- \\
\scratch OneFormer3D~\cite{kolodiazhnyi2024oneformer3d} &2024 &76.6 &- \\
\scratch ODIN~\cite{jain2024odin} &2024 &77.8 &74.4 \\
\scratch SparseUNet~\cite{choy20194d} &2019 &72.2 &73.6 \\
\ \ \pretrain PC~\cite{xie2020pointcontrast} &2020 &74.1 &- \\
\ \ \pretrain CSC~\cite{hou2021csc} &2021 &73.8 &- \\
\ \ \pretrain MSC~\cite{wu2023msc} &2023 &75.5 &- \\
\ \ \pretrain PPT (sup.)~\cite{wu2024ppt} &2023 &76.4 &76.6 \\
\scratch PTv1~\cite{zhao2021point} &2021 &70.6 &- \\
\scratch PTv2~\cite{wu2022ptv2} &2022 &75.4 &74.2 \\
\scratch PTv3~\cite{wu2024ptv3} &2023 &77.5 &77.9 \\
\ \ \pretrain MSC~\cite{wu2023msc} &2023 &78.2 &- \\
\ \ \pretrain PPT (sup.)~\cite{wu2024ppt} &2023 &78.6 &\cellcolor[HTML]{def3e6}\textbf{79.4} \\
\cellcolor[HTML]{f3f7fc}\ \ \pretrain Sonata~(linear probing) &2024 &72.5 &- \\
\cellcolor[HTML]{f3f7fc}\ \ \pretrain Sonata~(decoder probing) &2024 &79.1 &- \\
\cellcolor[HTML]{f3f7fc}\ \ \pretrain Sonata~(full fine-tuning) &2024 &\cellcolor[HTML]{def3e6}\textbf{79.4} &- \\

\bottomrule
\end{tabular}

%% file: table/appendix_s3dis.tex
\begin{tabular}{lccc}\toprule
Methods &Year &Area5 &6-fold \\\midrule
\scratch PointNet~\cite{qi2017pointnet} &2017 &41.1 &47.6\\
\scratch SegCloud~\cite{tchapmi2017segcloud} &2017 &48.9 &-\\
\scratch TanConv~\cite{tatarchenko2018tangent} &2018 &52.6 &-\\
\scratch PointCNN~\cite{li2018pointcnn} &2018 &57.3 &65.4\\
\scratch ParamConv~\cite{wang2018deep} &2018 &58.3 &-\\
\scratch PointWeb~\cite{zhao2019pointweb} &2019 &60.3 &66.7\\
\scratch HPEIN~\cite{jiang2019hierarchical} &2019 &61.9 &-\\
\scratch KPConv~\cite{thomas2019kpconv} &2019 &67.1 &70.6\\
\scratch GACNet~\cite{wang2019graph} &2019 &62.9 &-\\
\scratch PAT~\cite{yang2019modeling} &2019 &60.1 &-\\
\scratch SPGraph~\cite{landrieu2018large} &2018 &58.0 &62.1\\
\scratch SegGCN~\cite{lei2020seggcn} &2020 &63.6 &-\\
\scratch PAConv~\cite{xu2021paconv} &2021 &66.6 &-\\
\scratch StratifiedTransformer~\cite{lai2022stratified} &2022 &72.0 &- \\
\scratch PointNeXt~\cite{qian2022pointnext} &2022 &70.5 &74.9 \\
\scratch SuperpointTransformer~\cite{robert2023spt} & 2023 &68.9 &76.0 \\
\scratch PointMetaBase~\cite{lin2023pointmetabase} & 2023 &72.0 &77.0 \\
\scratch Swin3D~\cite{yang2023swin3d} &2023 &72.5 &76.9 \\
\ \ \pretrain Supervised~\cite{yang2023swin3d} &2023 &74.5 &79.8 \\
\scratch MinkUNet~\cite{choy20194d} &2019 &65.4 &65.4 \\
\ \ \pretrain PC~\cite{xie2020pointcontrast} &2020 &70.3 &- \\
\ \ \pretrain CSC~\cite{hou2021csc} &2021 &72.2 &- \\
\ \ \pretrain MSC~\cite{wu2023msc} &2023 &70.1 &- \\
\ \ \pretrain PPT (sup.)~\cite{wu2024ppt} &2023 &74.7 &78.1 \\
\scratch PTv1~\cite{zhao2021point} &2021 &70.4 &65.4 \\
\scratch PTv2~\cite{wu2022ptv2} &2022 &71.6 &73.5 \\
\scratch PTv3~\cite{wu2024ptv3}) &2023 &73.4 &77.7 \\
\ \ \pretrain PPT~\cite{wu2024ppt} &2023 &74.7 &80.8 \\
\cellcolor[HTML]{f3f7fc}\ \ \pretrain Sonata~(linear probing) &2024 &72.3 &76.5 \\
\cellcolor[HTML]{f3f7fc}\ \ \pretrain Sonata~(decoder probing) &2024 &74.5 &81.5 \\
\cellcolor[HTML]{f3f7fc}\ \ \pretrain Sonata~(full fine-tuning) &2024 &\cellcolor[HTML]{def3e6}\textbf{76.0} &\cellcolor[HTML]{def3e6}\textbf{82.3} \\
\bottomrule
\end{tabular}

%% file: main.bbl
\begin{thebibliography}{112}
\providecommand{\natexlab}[1]{#1}
\providecommand{\url}[1]{\texttt{#1}}
\expandafter\ifx\csname urlstyle\endcsname\relax
  \providecommand{\doi}[1]{doi: #1}\else
  \providecommand{\doi}{doi: \begingroup \urlstyle{rm}\Url}\fi

\bibitem[Armeni et~al.(2016)Armeni, Sener, Zamir, Jiang, Brilakis, Fischer, and Savarese]{armeni2016s3dis}
Iro Armeni, Ozan Sener, Amir~R. Zamir, Helen Jiang, Ioannis Brilakis, Martin Fischer, and Silvio Savarese.
\newblock 3d semantic parsing of large-scale indoor spaces.
\newblock In \emph{CVPR}, 2016.

\bibitem[Assran et~al.(2023)Assran, Duval, Misra, Bojanowski, Vincent, Rabbat, LeCun, and Ballas]{assran2023jpea}
Mahmoud Assran, Quentin Duval, Ishan Misra, Piotr Bojanowski, Pascal Vincent, Michael Rabbat, Yann LeCun, and Nicolas Ballas.
\newblock Self-supervised learning from images with a joint-embedding predictive architecture.
\newblock In \emph{CVPR}, 2023.

\bibitem[Avetisyan et~al.(2024)Avetisyan, Xie, Howard-Jenkins, Yang, Aroudj, Patra, Zhang, Frost, Holland, Orme, et~al.]{avetisyan2024scenescript}
Armen Avetisyan, Christopher Xie, Henry Howard-Jenkins, Tsun-Yi Yang, Samir Aroudj, Suvam Patra, Fuyang Zhang, Duncan Frost, Luke Holland, Campbell Orme, et~al.
\newblock Scenescript: Reconstructing scenes with an autoregressive structured language model.
\newblock In \emph{ECCV}, 2024.

\bibitem[Ba et~al.(2016)Ba, Kiros, and Hinton]{ba2016layer}
Jimmy~Lei Ba, Jamie~Ryan Kiros, and Geoffrey~E Hinton.
\newblock Layer normalization.
\newblock \emph{Stat}, 2016.

\bibitem[Baruch et~al.(2021)Baruch, Chen, Dehghan, Dimry, Feigin, Fu, Gebauer, Joffe, Kurz, Schwartz, and Shulman]{dehghan2021arkitscenes}
Gilad Baruch, Zhuoyuan Chen, Afshin Dehghan, Tal Dimry, Yuri Feigin, Peter Fu, Thomas Gebauer, Brandon Joffe, Daniel Kurz, Arik Schwartz, and Elad Shulman.
\newblock {ARK}itscenes - a diverse real-world dataset for 3d indoor scene understanding using mobile {RGB}-d data.
\newblock In \emph{NeurIPSW}, 2021.

\bibitem[Behley et~al.(2019)Behley, Garbade, Milioto, Quenzel, Behnke, Stachniss, and Gall]{behley2019semantickitti}
Jens Behley, Martin Garbade, Andres Milioto, Jan Quenzel, Sven Behnke, Cyrill Stachniss, and Jurgen Gall.
\newblock Semantickitti: A dataset for semantic scene understanding of lidar sequences.
\newblock In \emph{ICCV}, 2019.

\bibitem[Bengio et~al.(2009)Bengio, Louradour, Collobert, and Weston]{bengio2009curriculum}
Yoshua Bengio, J{\'e}r{\^o}me Louradour, Ronan Collobert, and Jason Weston.
\newblock Curriculum learning.
\newblock In \emph{ICML}, 2009.

\bibitem[Bochkovskii et~al.(2024)Bochkovskii, Delaunoy, Germain, Santos, Zhou, Richter, and Koltun]{bochkovskii2024depth}
Aleksei Bochkovskii, Ama{\"e}l Delaunoy, Hugo Germain, Marcel Santos, Yichao Zhou, Stephan~R Richter, and Vladlen Koltun.
\newblock Depth pro: Sharp monocular metric depth in less than a second.
\newblock \emph{arXiv preprint arXiv:2410.02073}, 2024.

\bibitem[Caesar et~al.(2020)Caesar, Bankiti, Lang, Vora, Liong, Xu, Krishnan, Pan, Baldan, and Beijbom]{caesar2020nuscenes}
Holger Caesar, Varun Bankiti, Alex~H Lang, Sourabh Vora, Venice~Erin Liong, Qiang Xu, Anush Krishnan, Yu Pan, Giancarlo Baldan, and Oscar Beijbom.
\newblock nuscenes: A multimodal dataset for autonomous driving.
\newblock In \emph{CVPR}, 2020.

\bibitem[Caron et~al.(2018)Caron, Bojanowski, Joulin, and Douze]{caron2018deep}
Mathilde Caron, Piotr Bojanowski, Armand Joulin, and Matthijs Douze.
\newblock Deep clustering for unsupervised learning of visual features.
\newblock In \emph{ECCV}, 2018.

\bibitem[Caron et~al.(2020)Caron, Misra, Mairal, Goyal, Bojanowski, and Joulin]{caron2020swav}
Mathilde Caron, Ishan Misra, Julien Mairal, Priya Goyal, Piotr Bojanowski, and Armand Joulin.
\newblock Unsupervised learning of visual features by contrasting cluster assignments.
\newblock In \emph{NeurIPS}, 2020.

\bibitem[Caron et~al.(2021)Caron, Touvron, Misra, J{\'e}gou, Mairal, Bojanowski, and Joulin]{caron2021emerging}
Mathilde Caron, Hugo Touvron, Ishan Misra, Herv{\'e} J{\'e}gou, Julien Mairal, Piotr Bojanowski, and Armand Joulin.
\newblock Emerging properties in self-supervised vision transformers.
\newblock In \emph{CVPR}, 2021.

\bibitem[Chen et~al.(2020)Chen, Kornblith, Norouzi, and Hinton]{chen2020simclr}
Ting Chen, Simon Kornblith, Mohammad Norouzi, and Geoffrey Hinton.
\newblock A simple framework for contrastive learning of visual representations.
\newblock In \emph{ICML}, 2020.

\bibitem[Chen et~al.(2024)Chen, Huang, Liu, Shen, Zhao, and Zhao]{chen2024anydoor}
Xi Chen, Lianghua Huang, Yu Liu, Yujun Shen, Deli Zhao, and Hengshuang Zhao.
\newblock Anydoor: Zero-shot object-level image customization.
\newblock In \emph{CVPR}, 2024.

\bibitem[Chen et~al.(2023)Chen, Liu, Zhang, Qi, and Jia]{chen2023largekernel3d}
Yukang Chen, Jianhui Liu, Xiangyu Zhang, Xiaojuan Qi, and Jiaya Jia.
\newblock Largekernel3d: Scaling up kernels in 3d sparse cnns.
\newblock In \emph{CVPR}, 2023.

\bibitem[Chiang et~al.(2019)Chiang, Lin, Liu, and Hsu]{chiang2019unified}
Hung-Yueh Chiang, Yen-Liang Lin, Yueh-Cheng Liu, and Winston~H Hsu.
\newblock A unified point-based framework for 3d segmentation.
\newblock In \emph{3DV}, 2019.

\bibitem[Choy et~al.(2019{\natexlab{a}})Choy, Gwak, and Savarese]{choy20194d}
Christopher Choy, JunYoung Gwak, and Silvio Savarese.
\newblock 4d spatio-temporal convnets: Minkowski convolutional neural networks.
\newblock In \emph{CVPR}, 2019{\natexlab{a}}.

\bibitem[Choy et~al.(2019{\natexlab{b}})Choy, Park, and Koltun]{choy2019fully}
Christopher Choy, Jaesik Park, and Vladlen Koltun.
\newblock Fully convolutional geometric features.
\newblock In \emph{ICCV}, 2019{\natexlab{b}}.

\bibitem[Contributors(2023)]{pointcept2023}
Pointcept Contributors.
\newblock Pointcept: A codebase for point cloud perception research.
\newblock \url{https://github.com/Pointcept/Pointcept}, 2023.

\bibitem[Contributors(2022)]{spconv2022}
Spconv Contributors.
\newblock Spconv: Spatially sparse convolution library.
\newblock \url{https://github.com/traveller59/spconv}, 2022.

\bibitem[Curless and Levoy(1996)]{curless1996volumetric}
Brian Curless and Marc Levoy.
\newblock A volumetric method for building complex models from range images.
\newblock In \emph{Proceedings of the 23rd annual conference on Computer graphics and interactive techniques}, pages 303--312, 1996.

\bibitem[Dai and Nie{\ss}ner(2018)]{dai20183dmv}
Angela Dai and Matthias Nie{\ss}ner.
\newblock 3dmv: Joint 3d-multi-view prediction for 3d semantic scene segmentation.
\newblock In \emph{ECCV}, 2018.

\bibitem[Dai et~al.(2017)Dai, Chang, Savva, Halber, Funkhouser, and Nie{\ss}ner]{dai2017scannet}
Angela Dai, Angel~X. Chang, Manolis Savva, Maciej Halber, Thomas Funkhouser, and Matthias Nie{\ss}ner.
\newblock Scannet: Richly-annotated 3d reconstructions of indoor scenes.
\newblock In \emph{CVPR}, 2017.

\bibitem[Darcet et~al.(2023)Darcet, Oquab, Mairal, and Bojanowski]{darcet2023vitneedreg}
Timothée Darcet, Maxime Oquab, Julien Mairal, and Piotr Bojanowski.
\newblock Vision transformers need registers.
\newblock \emph{arXiv:2309.16588}, 2023.

\bibitem[Doersch et~al.(2015)Doersch, Gupta, and Efros]{doersch2015unsupervised}
Carl Doersch, Abhinav Gupta, and Alexei~A Efros.
\newblock Unsupervised visual representation learning by context prediction.
\newblock In \emph{ICCV}, 2015.

\bibitem[Dosovitskiy et~al.(2015)Dosovitskiy, Fischer, Springenberg, Riedmiller, and Brox]{dosovitskiy2015discriminative}
Alexey Dosovitskiy, Philipp Fischer, Jost~Tobias Springenberg, Martin Riedmiller, and Thomas Brox.
\newblock Discriminative unsupervised feature learning with exemplar convolutional neural networks.
\newblock \emph{TPAMI}, 2015.

\bibitem[Engel et~al.(2023)Engel, Somasundaram, Goesele, Sun, Gamino, Turner, Talattof, Yuan, Souti, Meredith, et~al.]{engel2023aria}
Jakob Engel, Kiran Somasundaram, Michael Goesele, Albert Sun, Alexander Gamino, Andrew Turner, Arjang Talattof, Arnie Yuan, Bilal Souti, Brighid Meredith, et~al.
\newblock Project aria: A new tool for egocentric multi-modal ai research.
\newblock \emph{arXiv:2308.13561}, 2023.

\bibitem[Felzenszwalb and Huttenlocher(2004)]{felzenszwalb2004efficient}
Pedro~F Felzenszwalb and Daniel~P Huttenlocher.
\newblock Efficient graph-based image segmentation.
\newblock \emph{IJCV}, 2004.

\bibitem[Graham and van~der Maaten(2017)]{SubmanifoldSparseConvNet}
Benjamin Graham and Laurens van~der Maaten.
\newblock Submanifold sparse convolutional networks.
\newblock \emph{arXiv:1706.01307}, 2017.

\bibitem[Graham et~al.(2018)Graham, Engelcke, and van~der Maaten]{graham20183d}
Benjamin Graham, Martin Engelcke, and Laurens van~der Maaten.
\newblock 3d semantic segmentation with submanifold sparse convolutional networks.
\newblock In \emph{CVPR}, 2018.

\bibitem[Grill et~al.(2020)Grill, Strub, Altch{\'e}, Tallec, Richemond, Buchatskaya, Doersch, Avila~Pires, Guo, Gheshlaghi~Azar, et~al.]{grill2020byol}
Jean-Bastien Grill, Florian Strub, Florent Altch{\'e}, Corentin Tallec, Pierre Richemond, Elena Buchatskaya, Carl Doersch, Bernardo Avila~Pires, Zhaohan Guo, Mohammad Gheshlaghi~Azar, et~al.
\newblock Bootstrap your own latent-a new approach to self-supervised learning.
\newblock In \emph{NeruIPS}, 2020.

\bibitem[Gu et~al.(2023)Gu, Xiang, Li, Ling, Liu, Mu, Tang, Tao, Wei, Yao, Yuan, Xie, Huang, Chen, and Su]{gu2023maniskill2}
Jiayuan Gu, Fanbo Xiang, Xuanlin Li, Zhan Ling, Xiqiang Liu, Tongzhou Mu, Yihe Tang, Stone Tao, Xinyue Wei, Yunchao Yao, Xiaodi Yuan, Pengwei Xie, Zhiao Huang, Rui Chen, and Hao Su.
\newblock Maniskill2: A unified benchmark for generalizable manipulation skills.
\newblock In \emph{ICLR}, 2023.

\bibitem[Hariharan et~al.(2015)Hariharan, Arbelaez, Girshick, and Malik]{Hariharan_2015_CVPR}
Bharath Hariharan, Pablo Arbelaez, Ross Girshick, and Jitendra Malik.
\newblock Hypercolumns for object segmentation and fine-grained localization.
\newblock In \emph{CVPR}, 2015.

\bibitem[He et~al.(2020)He, Fan, Wu, Xie, and Girshick]{he2020moco}
Kaiming He, Haoqi Fan, Yuxin Wu, Saining Xie, and Ross Girshick.
\newblock Momentum contrast for unsupervised visual representation learning.
\newblock In \emph{CVPR}, 2020.

\bibitem[He et~al.(2022)He, Chen, Xie, Li, Doll{\'a}r, and Girshick]{he2022mae}
Kaiming He, Xinlei Chen, Saining Xie, Yanghao Li, Piotr Doll{\'a}r, and Ross Girshick.
\newblock Masked autoencoders are scalable vision learners.
\newblock In \emph{CVPR}, 2022.

\bibitem[Hjelm et~al.(2018)Hjelm, Fedorov, {Lavoie-Marchildon}, Grewal, Bachman, Trischler, and Bengio]{hjelm2018learning}
R.~Devon Hjelm, Alex Fedorov, Samuel {Lavoie-Marchildon}, Karan Grewal, Phil Bachman, Adam Trischler, and Yoshua Bengio.
\newblock Learning deep representations by mutual information estimation and maximization.
\newblock In \emph{ICLR}, 2018.

\bibitem[Hong et~al.(2023)Hong, Zhang, Gu, Bi, Zhou, Liu, Liu, Sunkavalli, Bui, and Tan]{hong2023lrm}
Yicong Hong, Kai Zhang, Jiuxiang Gu, Sai Bi, Yang Zhou, Difan Liu, Feng Liu, Kalyan Sunkavalli, Trung Bui, and Hao Tan.
\newblock Lrm: Large reconstruction model for single image to 3d.
\newblock \emph{arXiv preprint arXiv:2311.04400}, 2023.

\bibitem[Hou et~al.(2021)Hou, Graham, Nie{\ss}ner, and Xie]{hou2021csc}
Ji Hou, Benjamin Graham, Matthias Nie{\ss}ner, and Saining Xie.
\newblock Exploring data-efficient 3d scene understanding with contrastive scene contexts.
\newblock In \emph{CVPR}, 2021.

\bibitem[Hu et~al.(2020{\natexlab{a}})Hu, Yang, Xie, Rosa, Guo, Wang, Trigoni, and Markham]{hu2020randla}
Qingyong Hu, Bo Yang, Linhai Xie, Stefano Rosa, Yulan Guo, Zhihua Wang, Niki Trigoni, and Andrew Markham.
\newblock Randla-net: Efficient semantic segmentation of large-scale point clouds.
\newblock In \emph{CVPR}, 2020{\natexlab{a}}.

\bibitem[Hu et~al.(2020{\natexlab{b}})Hu, Zhen, Bai, Fu, and Tai]{hu2020jsenet}
Zeyu Hu, Mingmin Zhen, Xuyang Bai, Hongbo Fu, and Chiew-lan Tai.
\newblock Jsenet: Joint semantic segmentation and edge detection network for 3d point clouds.
\newblock In \emph{ECCV}, 2020{\natexlab{b}}.

\bibitem[Ioffe and Szegedy(2015)]{ioffe2015batch}
Sergey Ioffe and Christian Szegedy.
\newblock Batch normalization: Accelerating deep network training by reducing internal covariate shift.
\newblock In \emph{ICML}, 2015.

\bibitem[Jain et~al.(2024)Jain, Katara, Gkanatsios, Harley, Sarch, Aggarwal, Chaudhary, and Fragkiadaki]{jain2024odin}
Ayush Jain, Pushkal Katara, Nikolaos Gkanatsios, Adam~W. Harley, Gabriel Sarch, Kriti Aggarwal, Vishrav Chaudhary, and Katerina Fragkiadaki.
\newblock Odin: A single model for 2d and 3d perception.
\newblock In \emph{CVPR}, 2024.

\bibitem[James et~al.(2020)James, Ma, Rovick~Arrojo, and Davison]{james2019rlbench}
Stephen James, Zicong Ma, David Rovick~Arrojo, and Andrew~J. Davison.
\newblock Rlbench: The robot learning benchmark \& learning environment.
\newblock \emph{RAL}, 2020.

\bibitem[Jiang et~al.(2019)Jiang, Zhao, Liu, Shen, Fu, and Jia]{jiang2019hierarchical}
Li Jiang, Hengshuang Zhao, Shu Liu, Xiaoyong Shen, Chi-Wing Fu, and Jiaya Jia.
\newblock Hierarchical point-edge interaction network for point cloud semantic segmentation.
\newblock In \emph{ICCV}, 2019.

\bibitem[Jiang et~al.(2020)Jiang, Zhao, Shi, Liu, Fu, and Jia]{jiang2020pointgroup}
Li Jiang, Hengshuang Zhao, Shaoshuai Shi, Shu Liu, Chi-Wing Fu, and Jiaya Jia.
\newblock Pointgroup: Dual-set point grouping for 3d instance segmentation.
\newblock \emph{CVPR}, 2020.

\bibitem[Kolodiazhnyi et~al.(2024)Kolodiazhnyi, Vorontsova, Konushin, and Rukhovich]{kolodiazhnyi2024oneformer3d}
Maxim Kolodiazhnyi, Anna Vorontsova, Anton Konushin, and Danila Rukhovich.
\newblock Oneformer3d: One transformer for unified point cloud segmentation.
\newblock In \emph{CVPR}, 2024.

\bibitem[Lai et~al.(2022)Lai, Liu, Jiang, Wang, Zhao, Liu, Qi, and Jia]{lai2022stratified}
Xin Lai, Jianhui Liu, Li Jiang, Liwei Wang, Hengshuang Zhao, Shu Liu, Xiaojuan Qi, and Jiaya Jia.
\newblock Stratified transformer for 3d point cloud segmentation.
\newblock In \emph{CVPR}, 2022.

\bibitem[Landrieu and Simonovsky(2018)]{landrieu2018large}
Loic Landrieu and Martin Simonovsky.
\newblock Large-scale point cloud semantic segmentation with superpoint graphs.
\newblock In \emph{CVPR}, 2018.

\bibitem[Lei et~al.(2020)Lei, Akhtar, and Mian]{lei2020seggcn}
Huan Lei, Naveed Akhtar, and Ajmal Mian.
\newblock Seggcn: Efficient 3d point cloud segmentation with fuzzy spherical kernel.
\newblock In \emph{CVPR}, 2020.

\bibitem[Li et~al.(2018)Li, Bu, Sun, Wu, Di, and Chen]{li2018pointcnn}
Yangyan Li, Rui Bu, Mingchao Sun, Wei Wu, Xinhan Di, and Baoquan Chen.
\newblock Pointcnn: Convolution on x-transformed points.
\newblock \emph{NeurIPS}, 2018.

\bibitem[Liang et~al.(2024)Liang, Zhou, Xu, Zhu, Zou, Ye, Tan, and Bai]{liang2024pointmamba}
Dingkang Liang, Xin Zhou, Wei Xu, Xingkui Zhu, Zhikang Zou, Xiaoqing Ye, Xiao Tan, and Xiang Bai.
\newblock Pointmamba: A simple state space model for point cloud analysis.
\newblock In \emph{Advances in Neural Information Processing Systems}, 2024.

\bibitem[Lin et~al.(2023)Lin, Zheng, Li, Chao, Wang, Wang, Tian, and Ji]{lin2023pointmetabase}
Haojia Lin, Xiawu Zheng, Lijiang Li, Fei Chao, Shanshan Wang, Yan Wang, Yonghong Tian, and Rongrong Ji.
\newblock Meta architecture for point cloud analysis.
\newblock In \emph{CVPR}, pages 17682--17691, 2023.

\bibitem[Liu et~al.(2023)Liu, Shi, Kuang, Zhu, Li, Han, Cai, Porikli, and Su]{liu2023openshape}
Minghua Liu, Ruoxi Shi, Kaiming Kuang, Yinhao Zhu, Xuanlin Li, Shizhong Han, Hong Cai, Fatih Porikli, and Hao Su.
\newblock Openshape: Scaling up 3d shape representation towards open-world understanding.
\newblock In \emph{NeurIPS}, 2023.

\bibitem[Loshchilov(2019)]{loshchilov2019adamw}
I Loshchilov.
\newblock Decoupled weight decay regularization.
\newblock In \emph{ICLR}, 2019.

\bibitem[Loshchilov and Hutter(2017)]{loshchilov2017sgdr}
Ilya Loshchilov and Frank Hutter.
\newblock Sgdr: Stochastic gradient descent with warm restarts.
\newblock In \emph{ICLR}, 2017.

\bibitem[Ma et~al.(2022)Ma, Qin, You, Ran, and Fu]{ma2022rethinking}
Xu Ma, Can Qin, Haoxuan You, Haoxi Ran, and Yun Fu.
\newblock Rethinking network design and local geometry in point cloud: A simple residual mlp framework.
\newblock \emph{ICLR}, 2022.

\bibitem[Narita et~al.(2019)Narita, Seno, Ishikawa, and Kaji]{narita2019panopticfusion}
Gaku Narita, Takashi Seno, Tomoya Ishikawa, and Yohsuke Kaji.
\newblock Panopticfusion: Online volumetric semantic mapping at the level of stuff and things.
\newblock In \emph{IROS}, 2019.

\bibitem[Newcombe et~al.(2011)Newcombe, Izadi, Hilliges, Molyneaux, Kim, Davison, Kohi, Shotton, Hodges, and Fitzgibbon]{newcombe2011kinectfusion}
Richard~A Newcombe, Shahram Izadi, Otmar Hilliges, David Molyneaux, David Kim, Andrew~J Davison, Pushmeet Kohi, Jamie Shotton, Steve Hodges, and Andrew Fitzgibbon.
\newblock Kinectfusion: Real-time dense surface mapping and tracking.
\newblock In \emph{2011 10th IEEE international symposium on mixed and augmented reality}, pages 127--136. Ieee, 2011.

\bibitem[Noroozi and Favaro(2016)]{noroozi2016unsupervised}
Mehdi Noroozi and Paolo Favaro.
\newblock Unsupervised learning of visual representations by solving jigsaw puzzles.
\newblock In \emph{ECCV}, 2016.

\bibitem[Oquab et~al.(2024)Oquab, Darcet, Moutakanni, Vo, Szafraniec, Khalidov, Fernandez, Haziza, Massa, El-Nouby, Howes, Huang, Xu, Sharma, Li, Galuba, Rabbat, Assran, Ballas, Synnaeve, Misra, Jegou, Mairal, Labatut, Joulin, and Bojanowski]{oquab2023dinov2}
Maxime Oquab, Timothée Darcet, Theo Moutakanni, Huy~V. Vo, Marc Szafraniec, Vasil Khalidov, Pierre Fernandez, Daniel Haziza, Francisco Massa, Alaaeldin El-Nouby, Russell Howes, Po-Yao Huang, Hu Xu, Vasu Sharma, Shang-Wen Li, Wojciech Galuba, Mike Rabbat, Mido Assran, Nicolas Ballas, Gabriel Synnaeve, Ishan Misra, Herve Jegou, Julien Mairal, Patrick Labatut, Armand Joulin, and Piotr Bojanowski.
\newblock Dinov2: Learning robust visual features without supervision.
\newblock \emph{TMLR}, 2024.

\bibitem[Pang et~al.(2022)Pang, Wang, Tay, Liu, Tian, and Yuan]{Pang2022pointmae}
Yatian Pang, Wenxiao Wang, Francis~EH Tay, Wei Liu, Yonghong Tian, and Li Yuan.
\newblock Masked autoencoders for point cloud self-supervised learning.
\newblock In \emph{ECCV}, 2022.

\bibitem[Park et~al.(2022)Park, Jeong, Cho, and Park]{park2022fast}
Chunghyun Park, Yoonwoo Jeong, Minsu Cho, and Jaesik Park.
\newblock Fast point transformer.
\newblock In \emph{CVPR}, 2022.

\bibitem[Paszke et~al.(2019)Paszke, Gross, Massa, Lerer, Bradbury, Chanan, Killeen, Lin, Gimelshein, Antiga, et~al.]{paszke2019pytorch}
Adam Paszke, Sam Gross, Francisco Massa, Adam Lerer, James Bradbury, Gregory Chanan, Trevor Killeen, Zeming Lin, Natalia Gimelshein, Luca Antiga, et~al.
\newblock Pytorch: An imperative style, high-performance deep learning library.
\newblock \emph{NeurIPS}, 2019.

\bibitem[Peng et~al.(2024)Peng, Wu, Jiang, Chen, Zhao, Tian, and Jia]{peng2024oacnns}
Bohao Peng, Xiaoyang Wu, Li Jiang, Yukang Chen, Hengshuang Zhao, Zhuotao Tian, and Jiaya Jia.
\newblock Oa-cnns: Omni-adaptive sparse cnns for 3d semantic segmentation.
\newblock In \emph{CVPR}, 2024.

\bibitem[Qi et~al.(2017{\natexlab{a}})Qi, Su, Mo, and Guibas]{qi2017pointnet}
Charles~R Qi, Hao Su, Kaichun Mo, and Leonidas~J Guibas.
\newblock Pointnet: Deep learning on point sets for 3d classification and segmentation.
\newblock In \emph{CVPR}, 2017{\natexlab{a}}.

\bibitem[Qi et~al.(2017{\natexlab{b}})Qi, Yi, Su, and Guibas]{qi2017pointnet++}
Charles~R Qi, Li Yi, Hao Su, and Leonidas~J Guibas.
\newblock Pointnet++: Deep hierarchical feature learning on point sets in a metric space.
\newblock In \emph{NeurIPS}, 2017{\natexlab{b}}.

\bibitem[Qian et~al.(2022)Qian, Li, Peng, Mai, Hammoud, Elhoseiny, and Ghanem]{qian2022pointnext}
Guocheng Qian, Yuchen Li, Houwen Peng, Jinjie Mai, Hasan Hammoud, Mohamed Elhoseiny, and Bernard Ghanem.
\newblock Pointnext: Revisiting pointnet++ with improved training and scaling strategies.
\newblock In \emph{NeurIPS}, 2022.

\bibitem[Ramakrishnan et~al.(2021)Ramakrishnan, Gokaslan, Wijmans, Maksymets, Clegg, Turner, Undersander, Galuba, Westbury, Chang, Savva, Zhao, and Batra]{ramakrishnan2021hm3d}
Santhosh~Kumar Ramakrishnan, Aaron Gokaslan, Erik Wijmans, Oleksandr Maksymets, Alexander Clegg, John~M Turner, Eric Undersander, Wojciech Galuba, Andrew Westbury, Angel~X Chang, Manolis Savva, Yili Zhao, and Dhruv Batra.
\newblock Habitat-matterport 3d dataset ({HM}3d): 1000 large-scale 3d environments for embodied {AI}.
\newblock In \emph{NeurIPS}, 2021.

\bibitem[Robert et~al.(2023)Robert, Raguet, and Landrieu]{robert2023spt}
Damien Robert, Hugo Raguet, and Loic Landrieu.
\newblock Efficient 3d semantic segmentation with superpoint transformer.
\newblock In \emph{ICCV}, 2023.

\bibitem[Ronneberger et~al.(2015)Ronneberger, Fischer, and Brox]{ronneberger2015unet}
Olaf Ronneberger, Philipp Fischer, and Thomas Brox.
\newblock U-net: Convolutional networks for biomedical image segmentation.
\newblock In \emph{MICCAI}, 2015.

\bibitem[Rozenberszki et~al.(2022)Rozenberszki, Litany, and Dai]{rozenberszki2022scannet200}
David Rozenberszki, Or Litany, and Angela Dai.
\newblock Language-grounded indoor 3d semantic segmentation in the wild.
\newblock In \emph{ECCV}, 2022.

\bibitem[Sablayrolles et~al.(2019)Sablayrolles, Douze, Schmid, and J{\'e}gou]{sablayrolles2018spreading}
Alexandre Sablayrolles, Matthijs Douze, Cordelia Schmid, and Herv{\'e} J{\'e}gou.
\newblock Spreading vectors for similarity search.
\newblock In \emph{ICLR}, 2019.

\bibitem[Straub et~al.(2024)Straub, DeTone, Shen, Yang, Sweeney, and Newcombe]{straub2024efm3d}
Julian Straub, Daniel DeTone, Tianwei Shen, Nan Yang, Chris Sweeney, and Richard Newcombe.
\newblock Efm3d: A benchmark for measuring progress towards 3d egocentric foundation models.
\newblock \emph{arXiv:2406.10224}, 2024.

\bibitem[Sun et~al.(2020)Sun, Kretzschmar, Dotiwalla, Chouard, Patnaik, Tsui, Guo, Zhou, Chai, Caine, et~al.]{sun2020waymo}
Pei Sun, Henrik Kretzschmar, Xerxes Dotiwalla, Aurelien Chouard, Vijaysai Patnaik, Paul Tsui, James Guo, Yin Zhou, Yuning Chai, Benjamin Caine, et~al.
\newblock Scalability in perception for autonomous driving: Waymo open dataset.
\newblock In \emph{CVPR}, 2020.

\bibitem[Tatarchenko et~al.(2018)Tatarchenko, Park, Koltun, and Zhou]{tatarchenko2018tangent}
Maxim Tatarchenko, Jaesik Park, Vladlen Koltun, and Qian-Yi Zhou.
\newblock Tangent convolutions for dense prediction in 3d.
\newblock In \emph{CVPR}, 2018.

\bibitem[Tchapmi et~al.(2017)Tchapmi, Choy, Armeni, Gwak, and Savarese]{tchapmi2017segcloud}
Lyne Tchapmi, Christopher Choy, Iro Armeni, JunYoung Gwak, and Silvio Savarese.
\newblock Segcloud: Semantic segmentation of 3d point clouds.
\newblock In \emph{3DV}, 2017.

\bibitem[Thomas et~al.(2019)Thomas, Qi, Deschaud, Marcotegui, Goulette, and Guibas]{thomas2019kpconv}
Hugues Thomas, Charles~R Qi, Jean-Emmanuel Deschaud, Beatriz Marcotegui, Fran{\c{c}}ois Goulette, and Leonidas~J Guibas.
\newblock Kpconv: Flexible and deformable convolution for point clouds.
\newblock In \emph{ICCV}, 2019.

\bibitem[Thomas et~al.(2024)Thomas, Tsai, Barfoot, and Zhang]{thomas2024kpconvx}
Hugues Thomas, Yao-Hung~Hubert Tsai, Timothy~D Barfoot, and Jian Zhang.
\newblock Kpconvx: Modernizing kernel point convolution with kernel attention.
\newblock In \emph{CVPR}, 2024.

\bibitem[Vaswani et~al.(2017)Vaswani, Shazeer, Parmar, Uszkoreit, Jones, Gomez, Kaiser, and Polosukhin]{vaswani2017attention}
Ashish Vaswani, Noam Shazeer, Niki Parmar, Jakob Uszkoreit, Llion Jones, Aidan~N Gomez, {\L}ukasz Kaiser, and Illia Polosukhin.
\newblock Attention is all you need.
\newblock In \emph{NeurIPS}, 2017.

\bibitem[Wang et~al.(2024{\natexlab{a}})Wang, Jiang, Wu, Tian, Peng, Zhao, and Jia]{wang2024gc}
Chengyao Wang, Li Jiang, Xiaoyang Wu, Zhuotao Tian, Bohao Peng, Hengshuang Zhao, and Jiaya Jia.
\newblock Groupcontrast: Semantic-aware self-supervised representation learning for 3d understanding.
\newblock In \emph{CVPR}, 2024{\natexlab{a}}.

\bibitem[Wang et~al.(2019)Wang, Huang, Hou, Zhang, and Shan]{wang2019graph}
Lei Wang, Yuchun Huang, Yaolin Hou, Shenman Zhang, and Jie Shan.
\newblock Graph attention convolution for point cloud semantic segmentation.
\newblock In \emph{CVPR}, 2019.

\bibitem[Wang(2023)]{Wang2023OctFormer}
Peng-Shuai Wang.
\newblock Octformer: Octree-based transformers for {3D} point clouds.
\newblock \emph{SIGGRAPH}, 2023.

\bibitem[Wang et~al.(2018)Wang, Suo, Ma, Pokrovsky, and Urtasun]{wang2018deep}
Shenlong Wang, Simon Suo, Wei-Chiu Ma, Andrei Pokrovsky, and Raquel Urtasun.
\newblock Deep parametric continuous convolutional neural networks.
\newblock In \emph{CVPR}, 2018.

\bibitem[Wang et~al.(2024{\natexlab{b}})Wang, Leroy, Cabon, Chidlovskii, and Revaud]{wang2024dust3r}
Shuzhe Wang, Vincent Leroy, Yohann Cabon, Boris Chidlovskii, and Jerome Revaud.
\newblock Dust3r: Geometric 3d vision made easy.
\newblock In \emph{CVPR}, 2024{\natexlab{b}}.

\bibitem[Wu et~al.(2019)Wu, Qi, and Fuxin]{wu2019pointconv}
Wenxuan Wu, Zhongang Qi, and Li Fuxin.
\newblock Pointconv: Deep convolutional networks on 3d point clouds.
\newblock In \emph{CVPR}, 2019.

\bibitem[Wu et~al.(2023{\natexlab{a}})Wu, Fuxin, and Shan]{wu2023pointconvformer}
Wenxuan Wu, Li Fuxin, and Qi Shan.
\newblock Pointconvformer: Revenge of the point-based convolution.
\newblock In \emph{CVPR}, 2023{\natexlab{a}}.

\bibitem[Wu et~al.(2022)Wu, Lao, Jiang, Liu, and Zhao]{wu2022ptv2}
Xiaoyang Wu, Yixing Lao, Li Jiang, Xihui Liu, and Hengshuang Zhao.
\newblock Point transformer v2: Grouped vector attention and partition-based pooling.
\newblock In \emph{NeurIPS}, 2022.

\bibitem[Wu et~al.(2023{\natexlab{b}})Wu, Wen, Liu, and Zhao]{wu2023msc}
Xiaoyang Wu, Xin Wen, Xihui Liu, and Hengshuang Zhao.
\newblock Masked scene contrast: A scalable framework for unsupervised 3d representation learning.
\newblock In \emph{CVPR}, 2023{\natexlab{b}}.

\bibitem[Wu et~al.(2024{\natexlab{a}})Wu, Jiang, Wang, Liu, Liu, Qiao, Ouyang, He, and Zhao]{wu2024ptv3}
Xiaoyang Wu, Li Jiang, Peng-Shuai Wang, Zhijian Liu, Xihui Liu, Yu Qiao, Wanli Ouyang, Tong He, and Hengshuang Zhao.
\newblock Point transformer v3: Simpler, faster, stronger.
\newblock In \emph{CVPR}, 2024{\natexlab{a}}.

\bibitem[Wu et~al.(2024{\natexlab{b}})Wu, Tian, Wen, Peng, Liu, Yu, and Zhao]{wu2024ppt}
Xiaoyang Wu, Zhuotao Tian, Xin Wen, Bohao Peng, Xihui Liu, Kaicheng Yu, and Hengshuang Zhao.
\newblock Towards large-scale 3d representation learning with multi-dataset point prompt training.
\newblock In \emph{CVPR}, 2024{\natexlab{b}}.

\bibitem[Wu et~al.(2018)Wu, Xiong, Yu, and Lin]{wu2018unsupervised}
Zhirong Wu, Yuanjun Xiong, Stella~X. Yu, and Dahua Lin.
\newblock Unsupervised feature learning via non-parametric instance discrimination.
\newblock In \emph{CVPR}, 2018.

\bibitem[Xie et~al.(2021)Xie, Wang, Yu, Anandkumar, Alvarez, and Luo]{xie2021segformer}
Enze Xie, Wenhai Wang, Zhiding Yu, Anima Anandkumar, Jose~M Alvarez, and Ping Luo.
\newblock Segformer: Simple and efficient design for semantic segmentation with transformers.
\newblock In \emph{NeurIPS}, 2021.

\bibitem[Xie et~al.(2020)Xie, Gu, Guo, Qi, Guibas, and Litany]{xie2020pointcontrast}
Saining Xie, Jiatao Gu, Demi Guo, Charles~R Qi, Leonidas Guibas, and Or Litany.
\newblock Pointcontrast: Unsupervised pre-training for 3d point cloud understanding.
\newblock In \emph{ECCV}, 2020.

\bibitem[Xie et~al.(2022)Xie, Zhang, Cao, Lin, Bao, Yao, Dai, and Hu]{xie2022simmim}
Zhenda Xie, Zheng Zhang, Yue Cao, Yutong Lin, Jianmin Bao, Zhuliang Yao, Qi Dai, and Han Hu.
\newblock Simmim: A simple framework for masked image modeling.
\newblock In \emph{CVPR}, 2022.

\bibitem[Xu et~al.(2021)Xu, Ding, Zhao, and Qi]{xu2021paconv}
Mutian Xu, Runyu Ding, Hengshuang Zhao, and Xiaojuan Qi.
\newblock Paconv: Position adaptive convolution with dynamic kernel assembling on point clouds.
\newblock In \emph{CVPR}, 2021.

\bibitem[Yan et~al.(2020)Yan, Zheng, Li, Wang, and Cui]{yan2020pointasnl}
Xu Yan, Chaoda Zheng, Zhen Li, Sheng Wang, and Shuguang Cui.
\newblock Pointasnl: Robust point clouds processing using nonlocal neural networks with adaptive sampling.
\newblock In \emph{CVPR}, 2020.

\bibitem[Yang et~al.(2019)Yang, Zhang, Ni, Li, Liu, Zhou, and Tian]{yang2019modeling}
Jiancheng Yang, Qiang Zhang, Bingbing Ni, Linguo Li, Jinxian Liu, Mengdie Zhou, and Qi Tian.
\newblock Modeling point clouds with self-attention and gumbel subset sampling.
\newblock In \emph{CVPR}, 2019.

\bibitem[Yang et~al.(2024{\natexlab{a}})Yang, Kang, Huang, Xu, Feng, and Zhao]{yang2024depthanything}
Lihe Yang, Bingyi Kang, Zilong Huang, Xiaogang Xu, Jiashi Feng, and Hengshuang Zhao.
\newblock Depth anything: Unleashing the power of large-scale unlabeled data.
\newblock In \emph{CVPR}, 2024{\natexlab{a}}.

\bibitem[Yang et~al.(2024{\natexlab{b}})Yang, Kang, Huang, Zhao, Xu, Feng, and Zhao]{yang2024depthanythingv2}
Lihe Yang, Bingyi Kang, Zilong Huang, Zhen Zhao, Xiaogang Xu, Jiashi Feng, and Hengshuang Zhao.
\newblock Depth anything v2.
\newblock In \emph{NeurIPS}, 2024{\natexlab{b}}.

\bibitem[Yang et~al.(2023)Yang, Guo, Xiong, Liu, Pan, Wang, Tong, and Guo]{yang2023swin3d}
Yu-Qi Yang, Yu-Xiao Guo, Jian-Yu Xiong, Yang Liu, Hao Pan, Peng-Shuai Wang, Xin Tong, and Baining Guo.
\newblock Swin3d: A pretrained transformer backbone for 3d indoor scene understanding.
\newblock \emph{arXiv:2304.06906}, 2023.

\bibitem[Yeshwanth et~al.(2023)Yeshwanth, Liu, Nie{\ss}ner, and Dai]{yeshwanth2023scannet++}
Chandan Yeshwanth, Yueh-Cheng Liu, Matthias Nie{\ss}ner, and Angela Dai.
\newblock Scannet++: A high-fidelity dataset of 3d indoor scenes.
\newblock In \emph{ICCV}, 2023.

\bibitem[Yu et~al.(2022)Yu, Tang, Rao, Huang, Zhou, and Lu]{yu2021pointbert}
Xumin Yu, Lulu Tang, Yongming Rao, Tiejun Huang, Jie Zhou, and Jiwen Lu.
\newblock Point-bert: Pre-training 3d point cloud transformers with masked point modeling.
\newblock In \emph{CVPR}, 2022.

\bibitem[Zhang et~al.(2020)Zhang, Fang, Wah, and Torr]{zhang2020deep}
Feihu Zhang, Jin Fang, Benjamin Wah, and Philip Torr.
\newblock Deep fusionnet for point cloud semantic segmentation.
\newblock In \emph{ECCV}, 2020.

\bibitem[Zhang et~al.(2022{\natexlab{a}})Zhang, Li, Liu, Zhang, Su, Zhu, Ni, and Shum]{zhang2022dino}
Hao Zhang, Feng Li, Shilong Liu, Lei Zhang, Hang Su, Jun Zhu, Lionel~M. Ni, and Heung-Yeung Shum.
\newblock Dino: Detr with improved denoising anchor boxes for end-to-end object detection.
\newblock \emph{arXiv:2203.03605}, 2022{\natexlab{a}}.

\bibitem[Zhang et~al.(2022{\natexlab{b}})Zhang, Wang, Sagawa, Hashimoto, and Liang]{zhang2022fine}
Michael Zhang, Aditi~Raghunathan Wang, Shiori Sagawa, Tatsunori~B Hashimoto, and Percy Liang.
\newblock Fine-tuning can distort pretrained features and underperform out-of-distribution.
\newblock In \emph{ICLR}, 2022{\natexlab{b}}.

\bibitem[Zhang et~al.(2022{\natexlab{c}})Zhang, Guo, Gao, Fang, Zhao, Wang, Qiao, and Li]{zhang2022pointm2ae}
Renrui Zhang, Ziyu Guo, Peng Gao, Rongyao Fang, Bin Zhao, Dong Wang, Yu Qiao, and Hongsheng Li.
\newblock Point-m2ae: Multi-scale masked autoencoders for hierarchical point cloud pre-training.
\newblock In \emph{NeurIPS}, 2022{\natexlab{c}}.

\bibitem[Zhao et~al.(2019)Zhao, Jiang, Fu, and Jia]{zhao2019pointweb}
Hengshuang Zhao, Li Jiang, Chi-Wing Fu, and Jiaya Jia.
\newblock Pointweb: Enhancing local neighborhood features for point cloud processing.
\newblock In \emph{CVPR}, 2019.

\bibitem[Zhao et~al.(2021)Zhao, Jiang, Jia, Torr, and Koltun]{zhao2021point}
Hengshuang Zhao, Li Jiang, Jiaya Jia, Philip Torr, and Vladlen Koltun.
\newblock Point transformer.
\newblock In \emph{ICCV}, 2021.

\bibitem[Zheng et~al.(2020)Zheng, Zhang, Li, Tang, Gao, and Zhou]{zheng2020structured3d}
Jia Zheng, Junfei Zhang, Jing Li, Rui Tang, Shenghua Gao, and Zihan Zhou.
\newblock Structured3d: A large photo-realistic dataset for structured 3d modeling.
\newblock In \emph{ECCV}, 2020.

\bibitem[Zhou et~al.(2022{\natexlab{a}})Zhou, Wei, Wang, Shen, Xie, Yuille, and Kong]{zhou2021ibot}
Jinghao Zhou, Chen Wei, Huiyu Wang, Wei Shen, Cihang Xie, Alan Yuille, and Tao Kong.
\newblock ibot: Image bert pre-training with online tokenizer.
\newblock In \emph{ICLR}, 2022{\natexlab{a}}.

\bibitem[Zhou et~al.(2022{\natexlab{b}})Zhou, Wei, Wang, Shen, Xie, Yuille, and Kong]{zhou2022bert}
Jinghao Zhou, Chen Wei, Huiyu Wang, Wei Shen, Cihang Xie, Alan Yuille, and Tao Kong.
\newblock Image {BERT} pre-training with online tokenizer.
\newblock In \emph{ICLR}, 2022{\natexlab{b}}.

\bibitem[Zhou et~al.(2022{\natexlab{c}})Zhou, Koltun, and Kr{\"a}henb{\"u}hl]{zhou2022simple}
Xingyi Zhou, Vladlen Koltun, and Philipp Kr{\"a}henb{\"u}hl.
\newblock Simple multi-dataset detection.
\newblock In \emph{CVPR}, 2022{\natexlab{c}}.

\end{thebibliography}
